\documentclass[12pt]{article}
\usepackage{amsmath}
\usepackage{times}
\usepackage{graphicx}
\usepackage{color}
\usepackage{multirow}
\usepackage{lineno}
\usepackage{algorithm,caption}
\usepackage[noend]{algpseudocode}
\usepackage{amssymb}
\usepackage[onehalfspacing]{setspace}
\usepackage{changepage}
\usepackage[authoryear]{natbib}
\usepackage{rotating}
\usepackage{bbm}
\usepackage{latexsym}

\newcommand{\captionfonts}{\normalsize}

\makeatletter  
\long\def\@makecaption#1#2{%
  \vskip\abovecaptionskip
  \sbox\@tempboxa{{\captionfonts #1: #2}}%
  \ifdim \wd\@tempboxa >\hsize
    {\captionfonts #1: #2\par}
  \else
    \hbox to\hsize{\hfil\box\@tempboxa\hfil}%
  \fi
  \vskip\belowcaptionskip}
\makeatother   

\begin{document}
\hspace{13.9cm}1

\ \vspace{20mm}\\

{\LARGE Balancing New Against Old Information: The Role of Surprise in Learning}

\ \\
{\bf \large Mohammadjavad Faraji\,$^{\displaystyle 1}$, Kerstin Preuschoff\,$^{\displaystyle 2, \dagger}$, and Wulfram Gerstner\,$^{\displaystyle 1,\dagger,*}$}\\
{$^{\displaystyle 1}$School of Computer and Communication Sciences and School of Life Sciences, Brain Mind Institute, \'Ecole Polytechnique F\'ed\'eral de Lausanne, 1015 Lausanne EPFL, Switzerland.}\\
{$^{\displaystyle 2}$Geneva Finance Research Institute and Center for Affective Sciences, University of Geneva, 1211 Geneva, Switzerland.}\\
{$^{\dagger}$ Co-senior author}\\
{$*$ Corresponding author: wulfram.gerstner@epfl.ch}
%


\thispagestyle{empty}
\markboth{}{NC instructions}
\ \vspace{-0mm}\\
%


\begin{center} {\bf Abstract} \end{center}
Surprise describes a range of phenomena from unexpected events to behavioral responses. We propose a measure of surprise and use it for surprise-driven learning.
Our surprise measure takes into account data likelihood as well as the degree of commitment to a belief via the entropy of the belief distribution.  We find that surprise-minimizing learning dynamically adjusts the balance between new and old information without the need of knowledge about the temporal statistics of the environment. We apply our framework to a dynamic decision-making task and a maze exploration task. Our surprise minimizing framework  is suitable for learning in complex environments, even if the environment undergoes gradual or sudden changes and could eventually provide a framework to study the behavior of humans and animals encountering surprising events.


\section{Introduction}
To guide their behavior, humans and animals rely on previously learned knowledge about the world. Since the world is complex and models of the world are never perfect, the question arises whether we should trust our internal world model that we have built from past data or whether we should readjust it when we receive a new data sample. In noisy environments, a single  data sample may not be reliable and in general we need to average over  several data samples. However, when a structural change occurs in the environment, the most recent data samples are the most informative ones and we should put more weight on recent data samples than on earlier ones.

Indeed, both humans and animals adaptively adjust the relative contribution of old and newly acquired data during learning~\citep{behrens2007learning, nassar2012rational, krugel2009genetic, pearce1980model} and rapidly adapt to changing environments~\citep{pearce1980model, wilson1992restoration, holland1997brain}.
To capture this behaviour, existing models detect and respond to sudden changes using (absolute) reward prediction errors~\citep{hayden2011surprise, pearce1980model,roesch2012surprise}, risk prediction errors~\citep{preuschoff2007adding, preuschoff2008human}, uncertainty-based jump detection~\citep{nassar2010approximately, payzan2011risk} and hierarchical modeling~\citep{behrens2007learning, adams2007bayesian}.
Typically, in the above models, a low-dimensional variable, linked to the characteristics of the specific experimental design, is used to trigger a rebalancing between new and old information. 
Here we aim to generalize these approaches by using a more generic  `surprise' signal  as a trigger for shifting the balance between old and new information.

Quantities related to surprise have been previously used
in psychological theories of attention \citep{itti2009bayesian}, in statistical models of information theory \citep{shannon1948mathematical}
and in machine learning \citep{sun2011planning, frank2013curiosity, little2014learning, schmidhuber2010formal, singh2004intrinsically}.
For example, in artificial models of curiosity, surprise is linked to 
 sudden increases in information compression \citep{schmidhuber2010formal}.
Planning to be surprised so as to maximize information gain has been suggested as an optimal exploration technique in dynamic environments~\citep{sun2011planning, little2014learning}.
Maximizing an internally generated surprise signal
can drive active exploration for learning unknown environments, in the absence of external reward~\citep{frank2013curiosity}. In the framework of intrinsically motivated reinforcement learning \citep{singh2004intrinsically, oudeyer2007intrinsic}, researchers have defined ad-hoc features \citep{singh2004intrinsically, sutton2011horde, silver2016predictron} or information theoretic quantities~\citep{mohamed2015variational} that could replace the reward-prediction-error of classical reinforcement learning by a generalized model prediction error which could be surprise-related.
Furthermore, in the context of variational learning and free-energy minimization, a surprise measure
has been defined via the  model prediction error ~\citep{friston2010free, friston2009predictive, 10.3389/fncom.2014.00038, brea2013matching}.

Mathematically, human surprise is difficult to quantify~\citep{baldi2010bits, itti2009bayesian, palm2012novelty, tribus1961information, shannon1948mathematical}. Existing concepts can be roughly classified into two different categories. First, the log-likelihood of a single data point given a statistical model of the world has been called Shannon surprise or information content~\citep{shannon1948mathematical, tribus1961information, palm2012novelty}. Thus, in the context of these theories an unlikely event becomes a surprising event.
Information bottleneck approaches \citep{tishby2000information, mohamed2015variational} fall roughly into the same class.
Second, in the context of Baysian models of attention, surprise has been defined via the changes in model parameters induced by a new data point \citep{itti2009bayesian, baldi2010bits}. Thus, in these theories an event that causes a big change in the model of the world becomes a surprising event. Surprise as successful algorithmic compression of the agent's world model \citep{schmidhuber2010formal}  is  a non-Baysian formulation of a related idea. But what do we mean by (human) surprise?

The Webster dictionary defines surprise as ``an unexpected event, piece of information'' or ``the feeling caused by something that is unexpected or unusual'' [merriam-webster.com].
Note that `unexpected' is different from  `unlikely'.
An event can be unlikely without being unexpected: for example, you may park your car at the shopping mall next to a green BMW X3 with a license plate containing the number 5 without being surprised, even though the specific event is objectively very unlikely. But since you did not have any expectation, this specific event was not unexpected. A pure likelihood based definition of surprise, such as the Shannon information content~\citep{shannon1948mathematical, tribus1961information} cannot capture this aspect. Note that something can be {\it unexpected} only if the subject is committed to a belief about what to expect. As Matthew M. Hurley, Daniel C. Dennett and Reginald B. Adams have put it: `` what surprises us is ... things we expected \emph{not} to happen -- because we expected something else to happen instead. ''~\citep{hurley2011inside}. In other words, surprise arises from a mismatch between a strong opinion and a novel event, but this notion needs a more precise mathematical formulation.

In practice, humans know when they are surprised (egocentric view) indicating that there are specific physiological brain states corresponding to surprise.
Indeed, the state of surprise  in other humans (observer view)  is  detectable as 
startle responses~\citep{kalat2012biological} manifesting 
itself in  pupil dilation~\citep{hess1960pupil, preuschoff2011pupil} and tension in the muscles~\citep{kalat2012biological}. Neurally, the P300 component of the event-related potential~\citep{pineda1997human, missonnier1999automatic} measured by electroencephalography is associated with the violation of expectation~\citep{jaskowski1994suspense, kolossa2015computational}.
Furthermore,
surprising events have been shown to influence the development of the sensory cortex~\citep{fairhall2001efficiency}, and to drive attention~\citep{itti2009bayesian}, as well as learning and memory formation~\citep{ranganath2003neural, hasselmo1999neuromodulation, wallenstein1998hippocampus}.

As the first aim of this paper, we would like to develop a theory of surprise that captures the notion of unexpectedness in the sense of a mismatch between our current world model and the world model that the new data point implies. As a second aim of the paper we want to study how surprise can influence learning. We derive a class of learning rules that minimize the surprise if the same data point appears a second time. We will demonstrate in two examples why surprising events increase the speed of learning and show that surprise can be used as a trigger to balance new information against old one. 



\section{Results}

In the first subsection we introduce our notion of surprise and apply it to a few examples. In the second subsection we derive a learning rule from the principle of surprise minimization. The subsequent subsections apply this learning rule to two scenarios, starting with a one-dimensional prediction task, followed by a maze exploration corresponding to a parameter space with more than two hundred dimensions

\subsection{Definition of Surprise}

We aim for a measure of surprise which captures the notion of a mismatch between an opinion (current world model) and a novel event (data point) and which should have the following properties.

(i) surprise is different from statistical likelihood because it depends on the agent's commitment to her belief.

(ii) with the same level of commitment to a belief, surprise decreases with the likelihood of an event. 

(iii) for an event with a small likelihood, surprise increases with commitment to the belief. 

(iv) A surprising event will influence learning

While the final point will be the topic of the next subsection, we will present now a definition of surprise and check the properties (i) to (iii) by way of a few illustrative examples.

To mathematically formulate surprise, we assume that a subject receives data samples $X$ from an environment that is complex, potentially high-dimensional, only partially observable, stochastic, or changing over time. In contrast to an engineered environment where we might know the overall lay-out of the world (e.g., a hierarchical Markov decision process) and learn the unknown parameters from data,  we do not want to assume that we have knowledge about the lay-out of the world. Our world model may therefore be conceptually insufficient to capture the intrinsic structure  of the world and would therefore occasionally make wrong predictions even when we have observed large amounts of data. In short, our model of the world is expected to be simplistic and wrong - but since we know this we should be ready to readapt the world model when necessary. 

In our framework, we construct the world model from  many instances of simple models, each one characterized by a parameter $\theta \in \mathbb{R}^N$.
The likelihood of a data point $X$ under model $\theta$ is
$p(X|\theta)$. In a neuronal implementation, we may imagine that different instantiations of the model (with different parameter values $\theta$)
are represented in parallel by different (potentially overlapping) neuronal networks in the brain. If a new data point $X$ is provided as input to the sensory layer, the different models respond with activity $\hat{p}_X(\theta) = c \ p(X|\theta)$ with a suitable constant $c$ (which may depend on $X$ but is the same for all models).
The distribution $\hat{p}_X(\theta)$ represents the naive response of the whole brain network (i.e., of all models) in a setting where  all the models are equally likely. Formally, $\hat{p}_X(\theta)$ is the posterior probability under a flat prior (see {\bf Mathematical Methods}). We refer to $\hat{p}_X(\theta)$ as the ``scaled data likelihood'' of a naive observer.

However, not all models are equally likely. Based on the past observation of $n$ data points, the subject has formed an opinion which assigns to each model $\theta$ its relevance $\pi_n(\theta)$ for explaining the world. In a Bayesian framework, we could describe the likelihood of the new data point $X$ under the current opinion by $p(X) = \int_\theta p(X|\theta) \pi_n(\theta) d\theta$ where $\pi_n(\theta)$
summarizes the current opinion of the subject and the integral runs over all possible model instantiations, be it a finite number or a continuum. In case of a finite number $\theta_k, \ 1\leq k\leq K $, we can also think of $p(X)$ as the data likelihood in a mixture model with basis functions $p(X|\theta_k)$. However, since we are interested in surprise, we are not interested in 
the data likelihood but rather in  the degree of commitment of the subject to a specific opinion. The commitment is defined as the negative entropy of the current opinion:
\begin{equation}
  {\rm commitment} = -H(\pi_n) = \int_\theta  \pi_n(\theta) \ln \pi_n(\theta) 
  d\theta .
\end{equation}
A subject with a high commitment to her opinion (low entropy) will be viewed as a confident subject.

Surprise is the mismatch of a perceived data point $X_{n+1}$ with the current opinion. The current opinion (after observation of $n$ data samples $X_1,\dots, X_n$) is characterized by the distribution $\pi_n(\theta)$. The observed data point $X=X_{n+1}$ would lead in a naive observer to the scaled likelihood $\hat{p}_X(\theta)$ introduced above.
We define the surprise as the Kullback-Leibler divergence between these two distributions
\begin{equation}
S_{cc}(X;\pi_n) = D_{KL}[\pi_n(\theta)||\hat{p}_X(\theta)] = \int_\theta \pi_n(\theta) \ln \frac{\pi_n(\theta)}{\hat{p}_X(\theta)} \ d\theta.
\label{eq:AverageShannonModified}
\end{equation}

We call $S_{cc}$ a confidence corrected surprise because its definition includes the commitment to an opinion. To get acquainted with this definition, let us look at a few examples.

\begin{figure}[!h]
\centering
\includegraphics[height = 0.5\textwidth, width=1\textwidth]{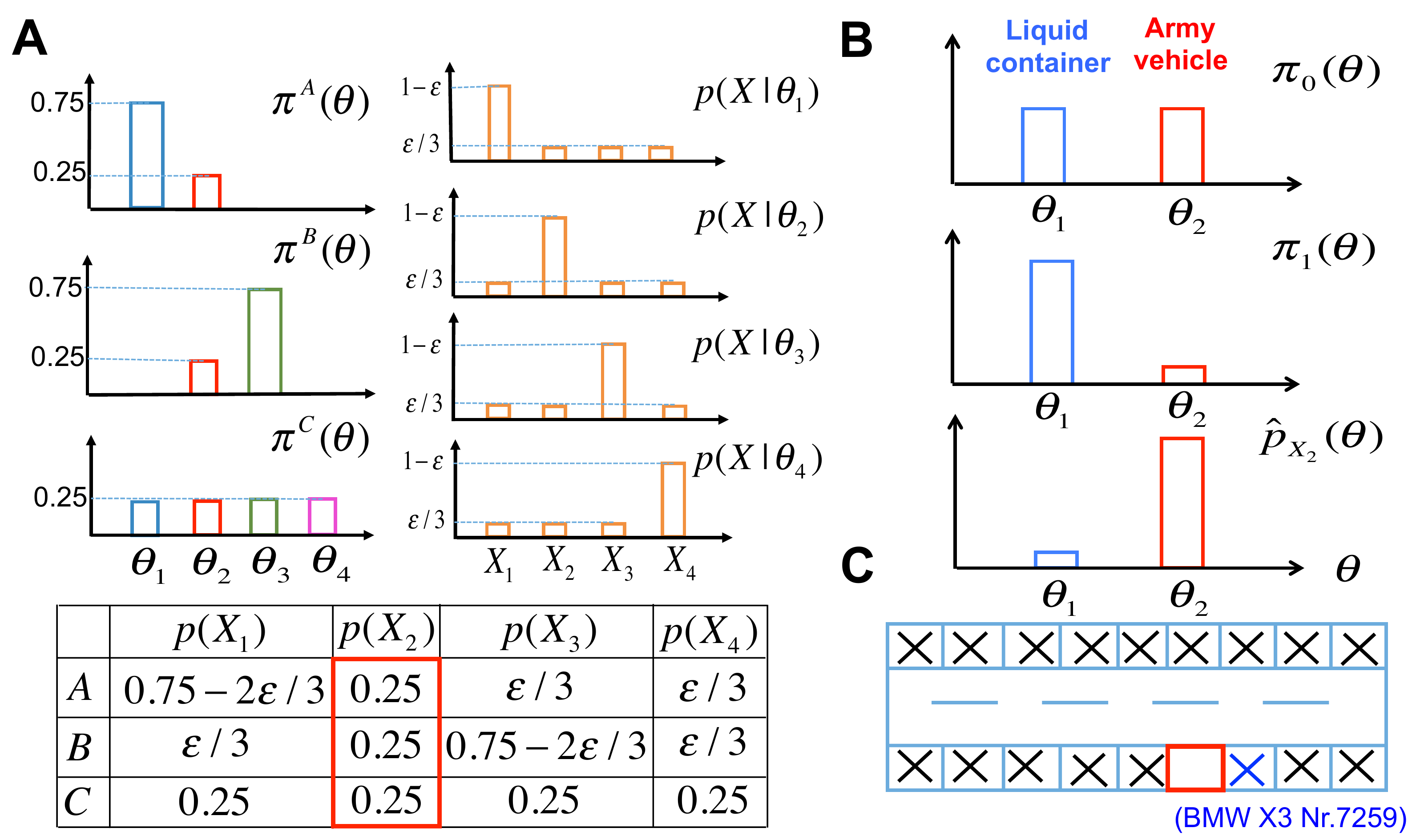}
\caption{{\bf Examples of surprise.}
  \textbf{A.} A committed person is more surprised than an uncommitted one.
  Three colleagues A, B and C have different believes (left) about the models $\theta_1, ..., \theta_4$ that determine the likelihood (right) that one of the four candidates is chosen as the CEO. Colleague A puts a lot of weight on the first model indicating his preference for the first candidate (top left). The table (bottom) indicates the likelihood of each candidate (columns) being chosen as the CEO (after marginalization over all models) for each colleague (rows). If the candidate number 2 is selected (column highlighted by red color), subject C is less surprised than subjects A and B because C is not committed to a specific opinion, although the likelihood of candidate 2 being chosen is considered the same ($0.25$) for all three colleagues.
    \textbf{B.} Surprise occurs only if a committed belief is disturbed. The first phrase of the goldfish-joke transforms our belief about the meaning ($\theta$) of  ``tank'' from $\pi_0(\theta)$ (top) to $\pi_1(\theta)$ (middle). The second phrase then causes in a naive observer a distribution
    $\hat{p}_{X_2}(\theta)$ (bottom) that is very different from the last belief $\pi_1(\theta)$ (middle); thus the listener is surprised.
  \textbf{C.} For a driver who just cares about having a free park slot, finding a park slot (empty red rectangle) next to a BMW X3 with the number plate 7259 (blue cross sign) is not very surprising, although it is very unlikely to occur.
  }
\label{fig:Fig 0}
\end{figure}

First, imagine that three colleagues (A, B, and C) wait for the outcome of the selection of the next CEO. Four candidates are in the running. Suppose that we have four models, $\theta_1, ..., \theta_4$ where model $\theta_k$  means candidate $X=k$ wins with probability $(1-\epsilon)$ (with small $\epsilon$) and the remaining probability is equally distributed amongst the other candidates. Formally, the model (or basis function) with parameter $\theta_k$ predicts outcome probabilities $p(X=k|\theta_k)=1-\epsilon$, and $p(X=k'|\theta_k)=\epsilon / 3$ for $k'\neq k$ (Fig.~\ref{fig:Fig 0}A, right). 

The current opinion $\{\pi^A(\theta), \pi^B(\theta), \pi^C(\theta)\}$ of each colleague about the four possible models corresponds to the histogram in Fig.~\ref{fig:Fig 0}A, left. Colleague A who is usually well informed has a weighting factor $\pi^A(\theta_1)=0.75$ for the first model, because he thinks the first candidate to be likely to win. According to his opinion, the first candidate wins with probability $p^A(X=1)=\sum_k p(X=1|\theta_k)\pi^A(\theta_k)$ and he gives lower probabilities to the other candidates (Fig.~\ref{fig:Fig 0}A, table). Colleague B has heard rumors and favors the third candidate while colleague C is uninformed as well as uninterested in the outcome and gives the same probabilities to each candidate. Note that colleagues A and B have the same commitment to their belief, i.e. $H(\pi^A)=H(\pi^B)$, but the likelihood of candidates differs. The commitment of colleague C is lower than that of A or B. 

Evaluation of the surprise measure indicates that (see {\bf Mathematical Methods} for the exact calculations) 

(a) if candidate 1 is selected, then A and B - despite having the same overall commitment to their belief - will be differently surprised due to different likelihoods of candidate 1 in their models.

(b) if candidate 2 is selected, then A is more surprised than if candidate 1 is selected because in his model candidate 2 is less likely.

(c) if candidate 2 is selected then A will be more surprised than C. Even though both colleagues assigned the same likelihood for this candidate, A's level of commitment to his belief is larger which leads to a bigger surprise.

Second, let us look at the theory of jokes developed by philosophers and cognitive psychologists~\citep{hurley2011inside} which emphasize that surprise in a joke can only work if the listener is committed to an opinion. Here is an example joke: ``There are two goldfish in a tank. One turns to the other and says: You man the guns, I'll drive''. The reason that some people find the  joke  funny is that ``a perception of the world (manning the guns and driving the tank) suddenly corrects our mistaken preconception (tank as a liquid container)'' \citep{hurley2011inside}.
Let us analyze the joke in the framework of our surprise measure.
A naive English speaking adult knows that tank can have two meanings, liquid container or military vehicle (Fig. \ref{fig:Fig 0}B, top). In the context of our theory, the two meanings correspond to two models (parameters $\theta_1$ and $\theta_2$ which have equal prior probability (opinion $\pi_0$).
In the first sentence of the joke, the word goldfish (data point $X_1$) shifts the belief of the listener to a situation where he gives more weight to the liquid container. This becomes the opinion $\pi_1$ of the listener (Fig. \ref{fig:Fig 0}B, middle). The opinion $\pi_1$ has low entropy, indicating a strong commitment.
Now comes the second sentence, with the words `driving' and `guns' which we may consider as data point $X_2$. These words trigger in a naive English speaking adult a distribution $\hat{p}_{X_2}(\theta)$ (Fig. \ref{fig:Fig 0}B, bottom)
which favors the interpretation of tank as a military vehicle.
Since the Kullback-Leibler divergence between the distributions in the second and third line is big, the listener is surprised.

Third, let us return to the example of the green BMW X3 with a 5 in the license plate, mentioned in the introduction. The likelihood of finding this type of car next to you in a shopping mall parking lot is extremely low (Fig. \ref{fig:Fig 0}C) yet you are not surprised. If it were the parking lot of a company where every morning you see a little red car on this very same parking slot, but today you see a green BMW you might be surprised - quite independent of the details of the green car. The difference arises from the degree of commitment.

The observations made in the above examples can be mathematically formalized as follows.

(1) Our measure of surprise as defined in
Eq.~(\ref{eq:AverageShannonModified})
is a linear combination of Shannon surprise and Bayesian surprise (and two further terms).
Because it contains Shannon surprise as one of the terms,  surprise decreases with increasing likelihood of the data under the current model
(see {\bf Mathematical Methods}).
This formal statement answers points (i) and (ii) from the beginning of the subsection.

(2) Our measure of surprise as defined in
Eq.~(\ref{eq:AverageShannonModified})
accounts for the differences in surprise between two subjects that reflect the differences in commitment to their opinion. In particular, a less-confident individual (lower commitment to the current opinion) will generally be less surprised than a confident individual who is strongly committed to her opinion (see {\bf Mathematical Methods}).
This formal statement answers point (iii) from the beginning of the subsection.

(3) Our measure of surprise as defined in
Eq.~(\ref{eq:AverageShannonModified})
can be computed rapidly because it only uses the scaled data likelihood of a naive observer 
and the degree of commitment to the current opinion. In particular, evaluation of surprise  needs neither
the lengthy evaluation of the posterior under the current model nor an update of the model parameters - in contrast to the Bayesian surprise model \citep{itti2005bayesian, baldi2010bits} with which our surprise measure  otherwise shares important properties
(see {\bf Mathematical Methods}).
The question of how  surprise can be used to influence learning is the topic of the next subsection.

\begin{figure}[!h]
\centering
\includegraphics[width=\textwidth]{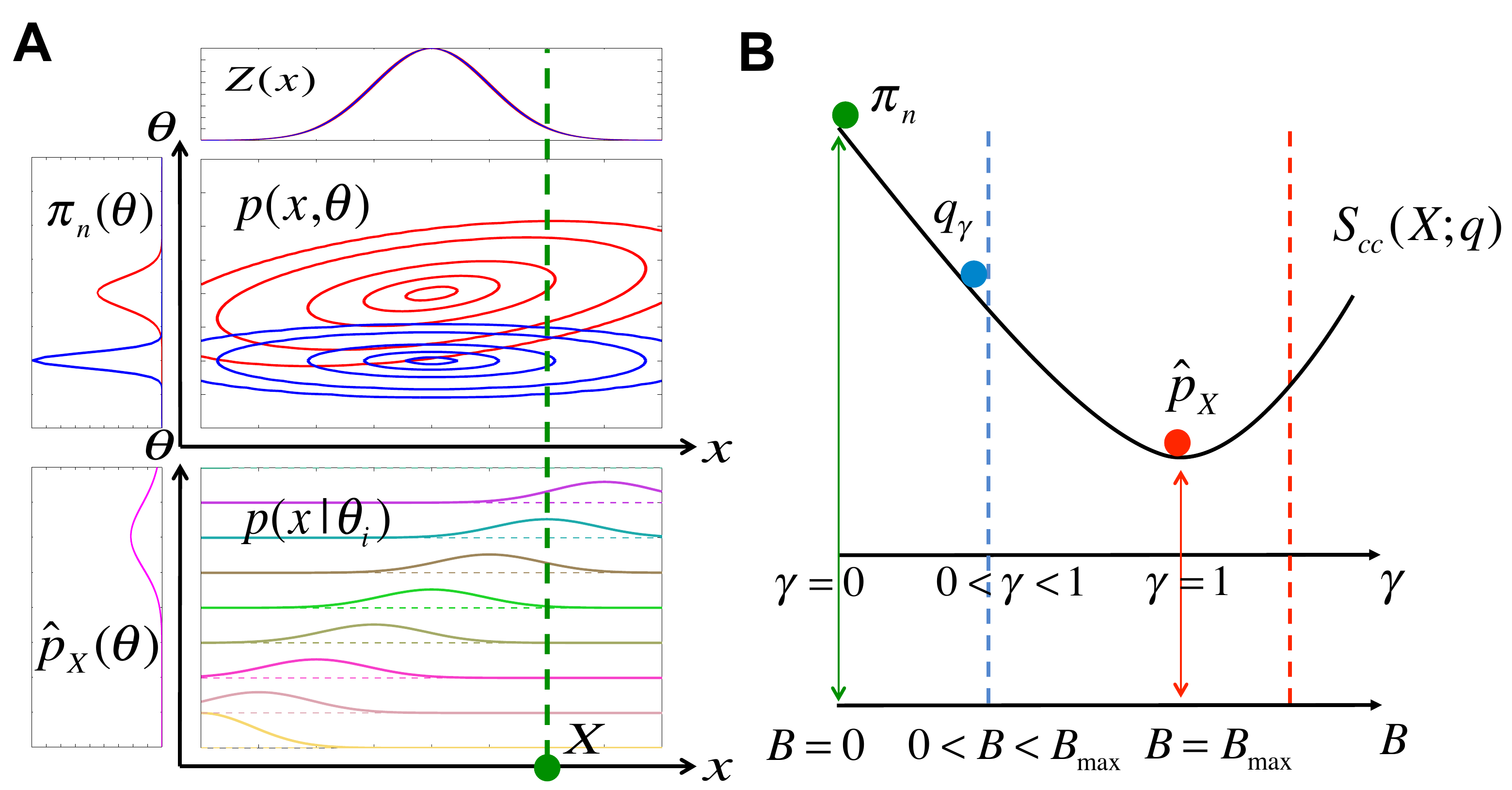}
\caption{{\bf Confidence-corrected surprise and constraint surprise minimization.} \textbf{A.} The impact of confidence on surprise. Top: Two distinct internal models (red and blue), described by joint distributions $p(x,\theta)$ (contour plots) over observable data $x$ and model parameters $\theta$, may have the same marginal distribution $Z(x)=\int_\theta p(x,\theta) d\theta$ (distributions along the $x$-axis coincide) but differ in the marginal distribution $\pi_n(\theta)=\int_x p(x,\theta) dx$ (distributions along the $\theta$-axis). Surprise measures that are computed with respect to $Z(x)$ neglect the uncertainty as measured by the entropy $H(\pi_n)$. Therefore, a given data sample $X$ (green dot) may be equally surprising in terms of the raw surprise $S_{raw}(X)$ [Eq~(\ref{eq:AverageShannon})] but results in higher confidence-corrected surprise $S_{cc}(X)$ [Eq~(\ref{eq:AverageShannonModified})] for the blue as compared to the red model, because $\pi_n$ in the red model is wider (corresponding to a larger entropy) than in the blue model. Bottom: The scaled likelihood $\hat{p}_X(\theta)$ (magenta) for the ``red'' internal model is calculated by evaluating the conditional probability distribution functions $p(x|\theta_i)$ (specified by different color for each $\theta_i$) at $x=X$ (intersection of dashed green line with colored curves). The confidence-corrected surprise $S_{cc}(X)$ is the KL divergence between $\hat{p}_X(\theta)$ (bottom, magenta) and $\pi_n(\theta)$ (top, red). \textbf{B.} Solutions to the (constraint) optimization problem in Eq~(\ref{eq:MinSurp}). The objective function, i.e. the updated value of the  surprise $S_{cc}(X;q)$ (black) for a given data sample $X$, is a parabolic landscape over $\gamma$ where each $\gamma$ corresponds to a unique belief distribution $q_{\gamma}$. Its global minimum is at $\gamma=1$ (corresponding to $q_1=\hat{p}_X$) which is equivalent to discarding all previously observed samples. The boundary $B$ constrains the range of $\gamma$ and thus the set of admissible belief distributions.
  At $B=0$ no change is allowed resulting in $\gamma=0$ with an updated belief equal
  to the current belief $\pi_n$ (green). $B\geq B_{max}=D_{KL}[\hat{p}_X||\pi_n]$ (red dashed line) implies that we allow to update the belief to a distribution further away from the current belief than the sample itself so the optimal solution is the scaled likelihood $\hat{p}_X$ or $\gamma =1$ as for the unconstrained problem. For $0<B<B_{max}$ (blue dashed line) the objective function is minimized by $q_\gamma$ in Eq~(\ref{eq:ModifiedBayesian2}) that fulfills the constraint $D_{KL}[q_\gamma||\pi_n]= B$ with $0<\gamma<1$.
}
\label{fig:Fig 1}
\end{figure}

We emphasize that our measure of surprise is not restricted to discrete models but can also be formulated for models with continuous parameters $\theta$ (see {\bf Mathematical Methods} - Fig. \ref{fig:Fig 1}A).
Our proposed measure of surprise is consistent with formulations of Schopenhauer 
that link surprise to the ``incongruity between representation of perception'' (in our framework: the scaled likelihood response $\hat{p}_X(\theta)$ to a data $X$) and ``abstract representations''  (in our framework:
the current opinion $\pi_n(\theta)$ formed from previous data points); freely cited after \citep{hurley2011inside}.

\subsection{Surprise minimization: the SMiLe-rule}

Successful learning implies an adaptation to the environment such that an event occurring for a second time is perceived as less surprising than the first time. In the following \emph{surprise minimization} refers to a learning strategy which modifies the internal model of the external world such that the unexpected observation becomes less surprising if it happens again in the near future. Surprise minimization is akin to -- though more general than -- reward prediction error learning. Reward based learning modifies the reward expectation such that a recurring reward results in a smaller reward prediction error. Similarly, surprise-minimization learning results in a smaller surprise for recurring events. 

To mathematically formulate learning through surprise minimization, we define a \emph{learning rule} $L(X,\pi_n)$ as a mapping from the current belief $\pi_n(\theta)$ to a new belief $q(\theta)$ after receiving data sample $X$, i.e., $q = L(X,\pi_n)$. Moreover, we define a \emph{belief update} as the learning step after a single data sample.

We define the class $\mathcal{L}$ of \emph{plausible} learning rules as the set of those learning rules $L$ for which the surprise $\mathcal{S}(X;q)$ of a given data sample $X$ under the new belief $q(\theta)$ is \emph{at most as surprising as} the surprise $\mathcal{S}(X;\pi_n)$ of that data sample under the current
belief $\pi_n(\theta)$, i.e.,

\begin{equation}
\mathcal{L} = \{L: \mathcal{S}(X;q) \leq \mathcal{S}(X;\pi_n), \ q=L(X,\pi_n), \forall X \in \mathcal{X}\}.
\label{eq:class}
\end{equation}

In other words, if the same data sample $X$ occurs a second time right after a belief update, it is perceived as less surprising than the first time.

After the belief update we have a new belief $\pi_{n+1}=q$ and we may ask how much the data $X$ has impacted the internal model. To answer this question we compare the surprise of data sample $X$ under the previous belief to its surprise under the new belief:

\begin{equation}
\Delta \mathcal{S}(X; L) = \mathcal{S}(X;\pi_n) - \mathcal{S}(X;\pi_{n+1}).
\label{eq:InfluenceFunctionVeryOriginal}
\end{equation}

Given a learning rule $L$, a data sample $X$ is considered more effective for a belief update than $X'$, if $\Delta \mathcal{S}(X; L) > \Delta \mathcal{S}(X'; L)$. Note that definitions in Eqs~(\ref{eq:class}) and (\ref{eq:InfluenceFunctionVeryOriginal}) do not depend on our specific choice of surprise measure $\mathcal{S}$. In the following we choose $\mathcal{S}$ to be the confidence-corrected surprise $S_{cc}$ [Eq~(\ref{eq:AverageShannonModified})].

The \emph{impact function} $\Delta S_{cc}(X;L)$ [Eq~(\ref{eq:InfluenceFunctionVeryOriginal})], for a given data sample $X$, is maximized by the learning rule that maps the new belief $\pi_{n+1}(\theta)$ to the scaled likelihood $\hat{p}_X(\theta)$. However, as the distribution $\pi_{n+1}=\hat{p}_X$ does not depend on the current
belief $\pi_n$, it discards all previously learned information. Therefore, it amounts to a valid though meaningless solution. 

To avoid overfitting to the last data sample, we need to limit our search to updated beliefs $q$ that are not too different from the current opinion $\pi_n$. This limited set can be expressed as the set of
new beliefs $q$ that fulfill the constraint $D_{KL}[q||\pi_n] \leq B$, for some non-negative upper bound $B \geq 0$. The parameter $B$ determines how much we allow our belief to change after receiving a data sample $X$. Maximizing the impact function $\Delta S_{cc}(X;L)$ under such a constraint, is equivalent to the following constraint optimization problem:

\begin{equation}
\underset{q:D_{KL}[q||\pi_n]\leq B}{\operatorname{min}} \ S_{cc}(X;q).
\label{eq:MinSurp}
\end{equation}

Using the method of Lagrange multipliers we find the solution of the minimization problem in Eq~(\ref{eq:MinSurp}) to be
\begin{equation}
q_\gamma(\theta) = \frac{p(X|\theta)^\gamma \pi_n(\theta)^{1-\gamma}}{Z(X;\gamma)},
\label{eq:ModifiedBayesian2}
\end{equation}
where $Z(X;\gamma)=\int_\theta p(X|\theta)^\gamma \pi_n(\theta)^{1-\gamma} \ d\theta$ is a normalizing factor and the parameter $\gamma$ with $0 \leq \gamma \leq 1$ is uniquely determined by the bound $B$ (see {\bf Mathematical Methods} for the proof). Moreover, for $\gamma<1$ the function $\gamma(B)$ increases monotonously. The unique relationship between $\gamma$ and $B$ means that once $B$ has been chosen, $\gamma$ is no longer a free parameter and vice versa.
Learning is implemented by using the solution of Eq.~(\ref{eq:ModifiedBayesian2}) as the new opinion:
$\pi_{n+1}(\theta) = q_\gamma(\theta)$.

Learning by updating according to
Eq~(\ref{eq:ModifiedBayesian2}) will be called \emph{surprise minimization learning (SMiLe)}
and we will  refer to Eq. (\ref{eq:ModifiedBayesian2}) as the SMiLe-rule.
The update step of the SMiLe-rule  is reminiscent of Bayes' rule except for the parameter $\gamma$ which modulates the relative contribution of the likelihood $p(X|\theta)$ and the current belief $\pi_n(\theta)$ to the
new belief $\pi_{n+1}(\theta) = q_\gamma(\theta)$. Note that the SMiLe rule belongs to the class $\mathcal{L}$ of plausible learning rules, for all $0 \leq \gamma\leq 1$.

Choosing $\gamma$ in the range $0 \leq \gamma \leq 1$ is equivalent to choosing a bound $B \geq 0$.
To understand how the optimal solution in Eq~(\ref{eq:ModifiedBayesian2}), and thus $\gamma$, relates to the boundary $B$, we illustrate its limiting cases (see Fig~\ref{fig:Fig 1}B): (i) $B=0$ yields $\gamma=0$ and the new belief $q$ is identical to the current belief $\pi_n$. In other words, the new information is discarded. (ii) For $B \geq B_{max}=D_{KL}[\hat{p}_X||\pi_n]$, the solution is always the scaled likelihood $\hat{p}_X$ (corresponding to $\gamma=1$) because $q=\hat{p}_X$ fulfills the constraint $D_{KL}[q||\pi_n]\leq B$ for any $B \geq B_{max}$ and minimizes $S_{cc}(X;q)$ among all possible belief distributions $q$. This is equivalent to the unconstrained case, and implies that all previous information is discarded. (iii) For $0 < B < B_{max}$ the optimal solution is the new belief $q_\gamma$ [Eq~(\ref{eq:ModifiedBayesian2})] with $0<\gamma <1$ satisfying $D_{KL}[q_\gamma||\pi_n]= B$. Moreover, $B>B'$ implies $\gamma>\gamma'$ (see Fig~\ref{fig:Fig 1}B, and {\bf Mathematical Methods} for the proof).

While the SMiLe rule [Eq~(\ref{eq:ModifiedBayesian2})] depends on a parameter $\gamma$ which is uniquely determined by the bound $B$, we have yet to indicate how to choose $B$. Highly surprising data should result in larger belief shifts. Therefore, the bound $B$ should increase with the level of surprise $S_{cc}$. 

The definition of an optimal (nonlinear) mapping from $S_{cc}$ to $B$ (and thus to $\gamma$) would require further assumptions about how surprise is related to the bound and we will therefore not search for optimality. However, it is instructive to study a few examples. For instance, if the nonlinear mapping were a step function, the system would make a binary choice between either keeping the old belief or relying on the last new data point. On the other hand, an extremely slow increase would amount to largely ignoring the surprise and sticking to the same old belief. Therefore, the sharpness of the transition in the mapping function matters. The \emph{exact} link between the bound and surprise is, however, not crucial as long as $B$ is monotonic in surprise in a \emph{reasonable} way. 

In the following, we choose a simple monotonic function to link the bound to the surprise. For each data sample $X$, we take 

\begin{equation}
B(X)=\frac{mS_{cc}(X;\pi_n)}{1+mS_{cc}(X;\pi_n)}B_{max}(X),
\label{eq:ChooseB}
\end{equation}
where $B_{max}(X)=D_{KL}[\hat{p}_X || \pi_n]$. Here, the monotonic function depends on a subject-specific parameter $m$ that describes an organism's propensity toward changing its belief. Note that in Eq~(\ref{eq:ChooseB}), $m=0$ indicates that the subject will never change her belief. As $m$ increases so does a subject's willingness to change her belief. We expect that differences in $m$ from one subject to the next will eventually allow us to capture heterogeneity in belief update strategies when fitting human behavior. Although $m$ is inserted in Eq~(\ref{eq:ChooseB}) to model subject dependence, one could also search for the best $m$ algorithmically in a given simulated environment or other computational setting.

Note that biological correlates of surprise such as pupil dilation or the activity of a neuromodulator will normally saturate at some maximal value, consistent with our choice of a saturating function in Eq~(\ref{eq:ChooseB}).

\subsection{Surprise-modulated belief update}

The surprise-modulated belief update combines the confidence-corrected surprise [Eq~(\ref{eq:AverageShannonModified})] and the SMiLe rule [Eq~(\ref{eq:ModifiedBayesian2})] to dynamically update our belief: after receiving a new data point $X$, we evaluate the surprise $S_{cc}(X;\pi_n)$ which sets the bound $B$ [Eq~(\ref{eq:ChooseB})] for our update and allows us to solve for $\gamma$. We then update the belief, using the SMiLe rule [Eq~(\ref{eq:ModifiedBayesian2})] with parameter $\gamma$ (see Algorithm~\ref{alg:modulated}).



\begin{algorithm}
\caption{Pseudo algorithm for surprise-modulated belief update (SMiLe)}\label{alg:modulated}
\begin{algorithmic}[1]
\State $N \gets$ number of data samples
\State ${\rm Belief} \gets \pi_0$ (the current belief) 
\State $m \gets 0.1$ (subject-dependent)
\For {$n$: $1$ to $N$}
\State $X_n \gets$ a new data sample
\State (i) evaluate the surprise $S_{cc}(X_n;{\rm Belief})$, Eq~(\ref{eq:AverageShannonModified})
\State (ii-a) calculate $B_{max}(X_n) = D_{KL}[\hat{p}_{X_n}||{\rm Belief}]$
\State (ii-b) choose the bound $B(X_n) = \frac{m S_{cc}(X_n;{\rm Belief})}{1+ m S_{cc}(X_n;{\rm Belief})} B_{max}(X_n)$
\State (iii) find $\gamma$ by solving $D_{KL}[q_\gamma||{\rm Belief}]=B(X_n)$
\State (iv) update using SMiLe, Eq~(\ref{eq:ModifiedBayesian2}): ${\rm Belief}(\theta) \gets \frac{p(X_n|\theta)^\gamma {\rm Belief}(\theta)^{1-\gamma}}{\int_\theta p(X_n|\theta)^\gamma {\rm Belief}(\theta)^{1-\gamma} \ d\theta}$ 
\EndFor
\State \textbf{Return} ${\rm Belief}$;
\Statex
\Statex \emph{Note 1}: In each iteration, we first calculate the surprise, step (i), before the model is updated in step (iv). 
\Statex \emph{Note 2}: The steps (ii-a), (ii-b), and (iii) can be merged and approximated by $\gamma = f(S_{cc}(X_n;{\rm Belief}))$ where $f(.)$ is a subjective function that increases with surprise.
\end{algorithmic}
\end{algorithm}

The parameter $\gamma$ in the SMiLe rule controls the \emph{impact} of a data sample $X$ on belief update such that a bigger $\gamma$ causes a larger impact. More precisely, the impact function $\Delta S_{cc}(X; L)$ in Eq~(\ref{eq:InfluenceFunctionVeryOriginal}), where $L$ is replaced by the SMiLe rule [Eq~(\ref{eq:ModifiedBayesian2})], is an increasing function of $\gamma$ (see {\bf Mathematical Methods} for the proof).

We note that in classical models of perception and attention~\citep{itti2009bayesian,baldi2010bits}, surprise has been defined as a measure of belief change such as $D_{KL} [\pi_{n+1}||\pi_n]$ or its mirror form $D_{KL}[\pi_n||\pi_{n+1}]$ where $\pi_{n+1}$ is calculated by Bayes formula, Eq~(\ref{eq:BayesRule}). We emphasize that our model of surprise is ``fast'' in the sense that it can be evaluated \emph{before} the beliefs are changed. Interestingly, the impact function is linked to the measure of \emph{belief change} by the following equation (see {\bf Mathematical Methods} for derivation),

\begin{equation}
\Delta S_{cc}(X; L) = \frac{1}{\gamma} D_{KL}[\pi_n||\pi_{n+1}] + \left( \frac{1}{\gamma} -1 \right) D_{KL}[\pi_{n+1} ||\pi_n] \geq 0,
\label{eq:InfluenceFunction}
\end{equation}
where $\pi_{n+1}=q$ is the new belief calculated with the SMiLe rule [Eq~(\ref{eq:ModifiedBayesian2})]. Therefore \emph{a larger reduction in the surprise implies a bigger change in belief}.

\subsection{Simulations}

In the following we will look at two examples to illustrate the functionality of our proposed surprise-modulated belief update Algorithm~\ref{alg:modulated}. The first one is a simple, one-dimensional dynamic decision-making task which has been used in behavioral studies~\citep{nassar2012rational, behrens2007learning} of learning under uncertainty. While somewhat artificial as a task, it is appealing as it nicely isolates different forms of uncertainty.
This allows us: (i) to demonstrate the basic quantities and properties of our algorithm, and
(ii) to show how its flexibility allows it to capture a wide range of behaviors. The second example is a multi-dimensional maze-exploration task which we will use to demonstrate how our algorithm extends to and performs in more complex and realistic experimental environments.

\subsubsection*{Gaussian estimation}

\begin{figure}[!h]
\centering
\includegraphics[height = 0.6\textwidth, width=0.9\textwidth]{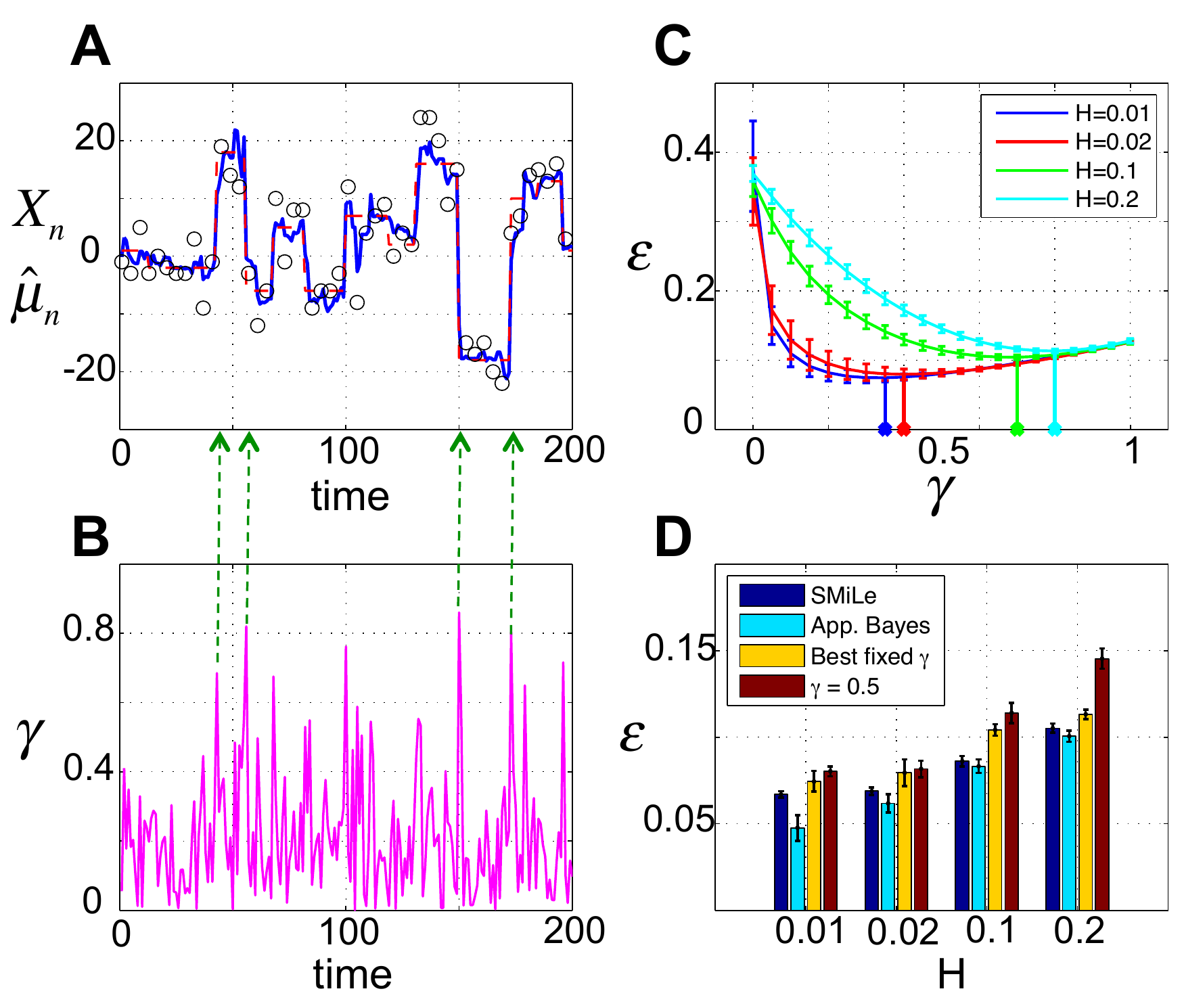}
\caption{{\bf Gaussian mean estimation task.} At each time step, a data sample $X_n$ is independently drawn from a normal distribution whose underlying mean may change within the interval $[-20,20]$ at unpredictable change points. On average, the underlying mean remains unchanged for $15$ time steps corresponding to a hazard rate $H=0.066$. The standard deviation of the distribution is fixed to $4$ and is assumed to be known to the subject. \textbf{A.} Using a surprise-modulated belief update (Algorithm~\ref{alg:modulated}), the estimated mean (blue) quickly approaches the true mean (dashed red) given observed samples (black circles). A few selected change points are indicated by green arrows. \textbf{B.} The weight factor $\gamma$ in Eq~(\ref{eq:SMiLe-Gaussian-task}) (magenta) increases at the change points, resulting in higher influence of newly acquired data samples on the new value of the mean. \textbf{C.} The estimation error $\epsilon$ per time step versus the weight factor $0 \leq \gamma \leq 1$ in the delta-rule method with constant $\gamma$ for four different hazard rates. The minimum estimation error (for best fixed $\gamma$) is achieved by a $\gamma$ (points on the horizontal axis) that decreases with the hazard rate, indicating that a bigger $\gamma$ is preferred in volatile environments. Error bars indicate standard deviation over all trials and $50$ episodes. \textbf{D.} For all models, the average estimation error $\epsilon$ increases with the hazard rate. Moreover, surprise-modulated belief update (SMile, dark blue) outperforms the delta-rule with the \emph{best} fixed $\gamma$ (Best fixed $\gamma$, yellow). The best fixed $\gamma$ for each hazard rate corresponds to the learning rate that has minimal estimation error (indicated by points on the horizontal axis in sub-figure \textbf{C}). Although the surprise-modulated SMile rule performs worse than the approximate Bayesian delta-rule~\citep{nassar2010approximately} (App. Bayes, light blue), the difference in the performance is not significant, except for the very small hazard rate of $0.01$.}
\label{fig:Fig 2}
\end{figure}

\textit{Task.} In the one-dimensional dynamic decision-making task, subjects are asked to estimate the mean of a distribution based on consecutively and independently drawn samples. At each time step $n$, a data sample $X_n$ is drawn from a normal distribution $\mathcal{N}(\mu_{n}, \sigma_{x}^2)$ and the subject is asked to provide her current estimate $\hat{\mu}_{n}$ of the mean of the distribution. Throughout the experiment, the mean may change without warning (Fig~\ref{fig:Fig 2}A). Changes occur with a \emph{hazard rate} of $H=0.066$. In Fig.~\ref{fig:Fig 2}C,~\ref{fig:Fig 2}D the hazard rate $H$ is varied. The variance $\sigma_x^2$ remains fixed.

\textit{Model.} We model the subject's belief \emph{before} the $n$-th sample $X_n$ is observed, as the normal distribution $\mathcal{N}(\hat{\mu}_{n-1}, \sigma_{n-1}^2)$ where $\hat{\mu}_{n-1}$ is the estimated mean and $\sigma_{n-1}^2$ determines how uncertain the subject is about her estimation. In order to keep the scenario as simple as possible, we assume $\sigma_0^2=\sigma_x^2$.

\textit{Results for the estimation task.} We find that the updated value of the mean $\hat{\mu}_n$ resulting from the surprise-modulated belief update (Algorithm \ref {alg:modulated}) is a \emph{weighted average} of the current estimate of the  mean $\hat{\mu}_{n-1}$ and the new sample $X_n$ (see {\bf Mathematical Methods} for derivation), 

\begin{equation}
\hat{\mu}_n=\gamma X_n + (1-\gamma)\hat{\mu}_{n-1}.
\label{eq:delta-rule}
\end{equation}

The weight factor, that determines to what extent a new sample $X_n$ affects the new mean $\hat{\mu}_n$, is determined by $\gamma$ which increases with the surprise $S_{cc}(X_n)$ of that sample (Fig~\ref{fig:Fig 2}B), i.e., 

\begin{equation}
\gamma = \sqrt{\frac{m S_{cc}(X_n)}{1 + m S_{cc}(X_n)}}, \quad S_{cc}(X_n) = \frac{(X_n-\hat{\mu}_{n-1})^2}{2\sigma_x^2}.
\label{eq:SMiLe-Gaussian-task}
\end{equation}

Note that in this example, the confidence-corrected surprise measure is related to the \emph{normalized unsigned prediction error} $|X_n-\hat{\mu}_{n-1}|/\sigma_x$. This outcome of our SMiLe-update is consistent with recent approaches in reward learning that suggest to use reward prediction errors scaled by standard deviation or variance~\citep{preuschoff2007adding}.

The confidence-corrected surprise increases suddenly in response to the samples immediately after the change points, as they are unexpected under the current belief. As a consequence, surprising samples increase the influence of a new data sample on the estimated mean (Fig~\ref{fig:Fig 2}B).

We compared our surprise modulated belief update [Eqs~(\ref{eq:delta-rule}) and (\ref{eq:SMiLe-Gaussian-task})] with a delta-rule [Eq~(\ref{eq:delta-rule})] with \emph{constant} weighting factor $\gamma$. To enable a fair comparison we consider two situations: (i) we arbitrarily fix $\gamma$ at $0.5$ or (ii) for a given hazard rate $H$, we first search for the optimal value of fixed $\gamma$ so as to minimize the estimation error (Fig~\ref{fig:Fig 2}C). We find that our surprise-modulated belief update outperforms the delta-rule with \emph{any} constant learning rate (Fig~\ref{fig:Fig 2}D). This clearly shows that an adaptive learning rate is preferable to a fixed learning rate. 

We also compared our proposed algorithm with a delta-rule that approximates the optimal Bayesian solution~\citep{nassar2010approximately}. In the optimal model, the subject knows a-priori that the mean will change at unknown points in time, i.e., the subject makes use of a hierarchical statistical model of the world. The algorithm proposed in~\citep{nassar2010approximately} provides an efficient approximate solution to estimate the parameters of the hierarchical model. In this algorithm, the subject increases the learning rate as a function of the probability of encountering a change point at a given time step. This probability requires knowledge or online estimation of the hazard rate, which indicates how frequently change points occur. Although our surprise-modulated belief update does not outperform the approximate Bayesian delta-rule, the difference in performance is, in most cases, not significant (see Fig~\ref{fig:Fig 2}D). In other words, our method, which does not require any information about the hazard rate, can nearly reach the quality of the optimal Bayesian solution, with significantly reduced computational complexity. Note that the SMiLe rule is not designed for (almost) stationary environments where no fundamental change in context occurs. Therefore, in the case where the true mean is constant (low hazard rate), the SMiLe rule results in increased volatility in estimation. This is why the difference in performance of SMiLe and the optimized Bayesian delta-rule becomes more evident for smaller hazard rates than bigger ones (see Fig~\ref{fig:Fig 2}D).

\subsubsection*{Maze exploration}

\begin{figure}[!h]
\centering
\includegraphics[height=0.6\textwidth, width=0.8\textwidth]{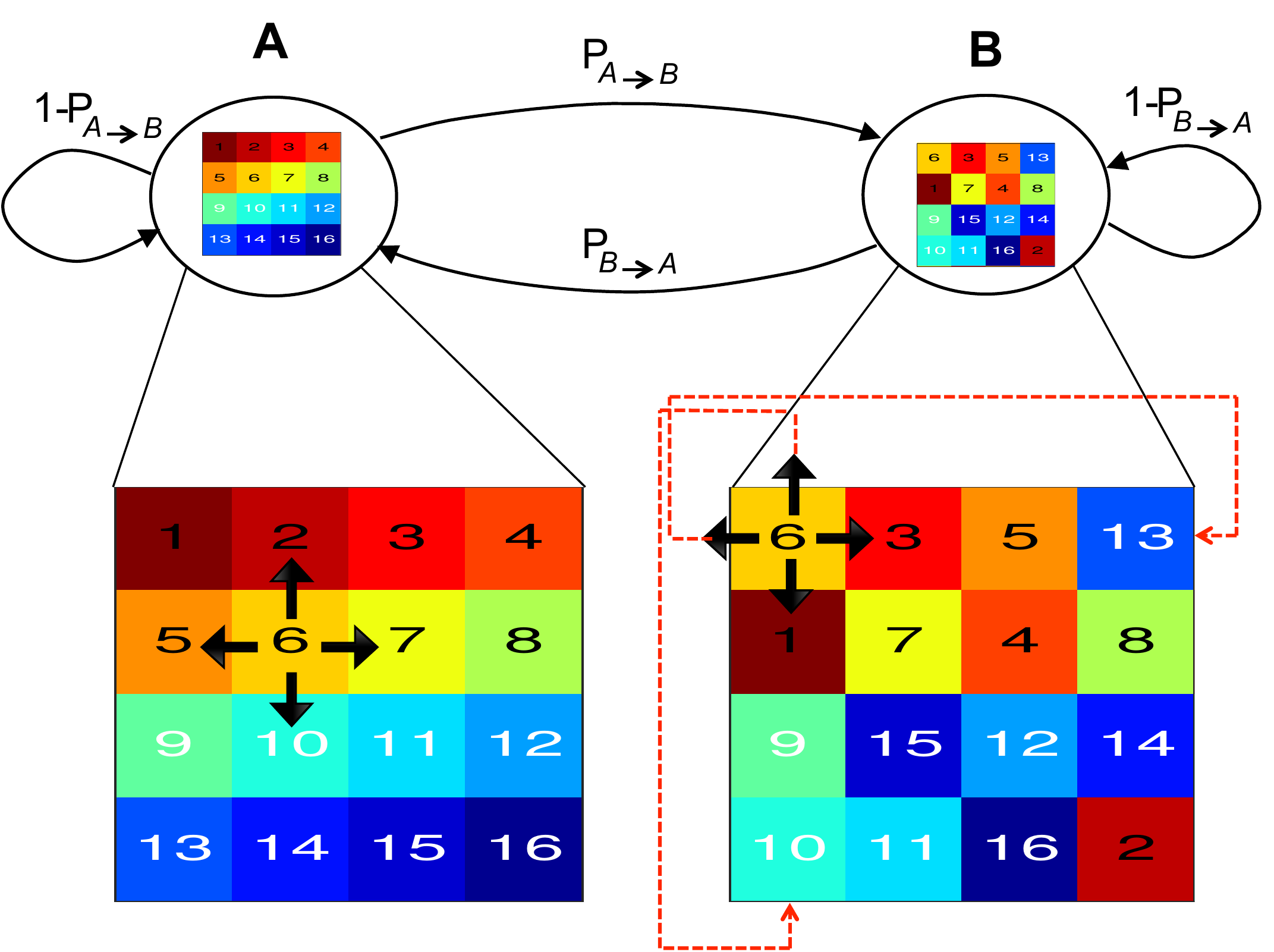}
\caption{{\bf Maze exploration task.} Environments $\mathcal{A}$ (left) and $\mathcal{B}$ (right) both consist of $16$ rooms, but differ in topology. At each time step, one of the four available doors (up, down, right, left) in the current room (e.g. $s=6$) is randomly opened (with probability $0.25$). While the learning agent is in environment $\mathcal{A}$, the environment may change to $\mathcal{B}$ with probability $P_{\mathcal{A}\rightarrow \mathcal{B}} \leq 0.1$ in the next time step of duration $\Delta t$. Similarly, $P_{\mathcal{B}\rightarrow \mathcal{A}}$ indicates the environment switches from $\mathcal{B}$ to $\mathcal{A}$. Therefore, as the agent starts moving out of state $s=6$, depending on the current environment and switch probabilities $P_{\mathcal{A}\rightarrow \mathcal{B}}$ and $P_{\mathcal{B}\rightarrow \mathcal{A}}$, it will end up in environment $\mathcal{A}$ (i.e., $s'\in\{2,10,7,5\}$) or $\mathcal{B}$ (i.e., $s'\in\{10,1,3,13\}$). The duration of a stay in environment $\mathcal{A}$ is therefore exponentially distributed with mean $\tau_\mathcal{A}=\Delta t / P_{\mathcal{A}\rightarrow \mathcal{B}}$, where the parameter $\tau_{\mathcal{A}}$ determines the \emph{time scale of stability} in environment $\mathcal{A}$, i.e., for larger $\tau_\mathcal{A}$ an agent has more time for adapting to $\mathcal{A}$ after a change point. The \emph{expected fraction of time spent in total} within environment $\mathcal{A}$ is equal to $\psi_{\mathcal{A}} = P_{\mathcal{B}\rightarrow \mathcal{A}} /(P_{\mathcal{B}\rightarrow \mathcal{A}} + P_{\mathcal{A}\rightarrow \mathcal{B}})$. Note that $\tau_\mathcal{A}$ and $\psi_\mathcal{A}$ are two free parameters that we can change to study how the agent performs in different circumstances (e.g., see Fig~\ref{fig:Fig 7}). }
\label{fig:Fig 3}
\end{figure}

\textit{Task.} The maze exploration task is similar to tasks used in behavioral neuroscience and robotics~\citep{morris1984developments, gillner1998navigation, nelson2004maze, 10.3389/fncom.2014.00038}.
There are two environments $\mathcal{A}$ and $\mathcal{B}$, each composed of the same uniquely labeled (e.g., by colors or cue cards) rooms. $\mathcal{A}$ and $\mathcal{B}$ only differ in the spatial arrangement (topology) of rooms (see Fig~\ref{fig:Fig 3}). Neighboring rooms are connected and accessible through doors. Initially, the agent is placed into either $\mathcal{A}$ or $\mathcal{B}$. At each time step, a door of the current room opens and the agent moves into the adjacent room, thus exploring the environment. After a random exploration time the environment is switched. The agent is not informed that a switch has occurred. Once the environment is changed, the agent must quickly adapt to the new environment. Note that this task differs from a reinforcement learning task because the task at hand just consists of the \emph{exploration} phase. In particular, there is no reward involved in learning. 

\textit{Model.} We model the knowledge of the environment by a learning agent that updates a set of parameters $\alpha(s,\check{s})\geq 1$ used for describing its belief about \emph{state transitions} from $s\in \{1,2,...,16\}$ to $\check{s} \in \{1,2,...,16\} \backslash s$, where $16$ is the number of rooms.
More precisely, an agent's belief about how likely it is to visit $\check{s}$, given the current state $s$, is modeled by a \emph{Dirichlet distribution} parametrized by a \emph{vector} of parameters $\vec{\alpha}(s) \in \mathbb{R}^{15}$. The components of the vector $\vec{\alpha}(s)$ are denoted as $\alpha(s,\check{s})$.
We emphasize that the agent has a structurally incomplete model of the world since it does not know that there are two different environments.

In order to see how well our proposed surprise-modulated belief update algorithm performs in this task, we compare it with a naive Bayesian learner and an online expectation-maximization (EM) algorithm~\citep{mongillo2008online}. While in the former the agent assumes that there is only a single stable, but stochastic environment, the latter benefits from knowing the true hidden Markov model (HMM) of the task and approximates the optimal hierarchical Bayesian solution (see {\bf Mathematical Methods}).

\textit{Results for the maze task.} The surprise-modulated belief update (Algorithm~\ref{alg:modulated}), with the Dirichlet distribution inserted, yields Algorithm~\ref{alg:modulated-maze} for the maze exploration task (see {\bf Mathematical Methods} for derivation). Immediately after a transition from the current state $s$ to the next state $s'$, the
new belief $q_\gamma$ obtained by the SMiLe rule [Eq~(\ref{eq:ModifiedBayesian2})] is a Dirichlet distribution $\vec{\alpha}_{new}(s)$ with components $\alpha_{new}(s,\check{s})=\gamma (1+[\check{s}=s']) + (1-\gamma) \alpha_{old}(s,\check{s})$, that can be written as a \emph{weighted average} of the parameters of the current belief $\pi_n$ (i.e., $\alpha_{old}(s,\check{s})$) and those of the scaled likelihood $\hat{p}_X$ (i.e., $1+[\check{s}=s']$). Here, $[\check{s}=s']$ indicates a number that is $1$ if the condition in square brackets is satisfied, and $0$ otherwise.


\begin{algorithm}
\caption{Surprise-modulated belief update for the maze exploration task}\label{alg:modulated-maze}
\begin{algorithmic}[1]
\State $N \gets$ number of data samples
\State $\alpha(s,\check{s}) = 1, \quad \forall s\in\{1,2,...,16\}, \check{s}\in\{1,2,...,16\} \backslash \{s\}$ (a uniform prior belief) 
\State $m \gets 0.1$ (subject-dependent)
\State Start in state $s$
\For {$n$: $1$ to $N$}
\Statex $\ \ \quad $ \# at this time step we only update the parameters that describe state transitions from the \underline{current state $s$} to all possible next states $\check{s}\in \{1,2,...,16\} \backslash \{s\}$. The current belief, for the state $s$, is $\pi_{n-1} \sim Dir(\mathbf{a}), \ \mathbf{a} \in \mathbb{R}^{15}, \ \mathbf{a}(\check{s})=\alpha (s,\check{s})$.
\State $X_n:s \rightarrow s'$ (a new transition is observed)
\Statex $\ \ \quad $ \# the scaled likelihood is $\hat{p}_X \sim Dir(\mathbf{b}), \mathbf{b} \in \mathbb{R}^{15}, \ \mathbf{b}(\check{s})=1+[\check{s}=s']$ 
\State (i) $S_{cc}(X_n;\pi_{n-1})= D_{KL}[Dir(\mathbf{a})||Dir(\mathbf{b})]$
\State (ii-a) $B_{max}(X_n) = D_{KL}[Dir(\mathbf{b})||Dir(\mathbf{a})]$
\State (ii-b) $B(X_n) = \frac{m S_{cc}(X_n;\pi_{n-1})}{1+ m S_{cc}(X_n;\pi_{n-1})} B_{max}(X_n)$
\State (iii) find $\gamma$ by solving $D_{KL}[Dir(\gamma \mathbf{b}+(1-\gamma)\mathbf{a})||Dir(\mathbf{a})]=B(X_n)$
\State (iv) $\alpha(s,\check{s}) \gets (1-\gamma) \alpha(s,\check{s}) + \gamma (1+[\check{s}=s'])$ 
\EndFor
\State \textbf{Return} $\alpha(s,\check{s}), \forall s,\check{s}$;
\Statex
\Statex \emph{Note 1}: $D_{KL}[Dir(\mathbf{m})||Dir(\mathbf{n})] = \ln \Gamma(\sum_{\check{s}} \mathbf{m}(\check{s})) - \ln \Gamma(\sum_{\check{s}} \mathbf{n}(\check{s})) - \sum_{\check{s}} \ln \Gamma(\mathbf{m}(\check{s})) + \sum_{\check{s}} \ln \Gamma(\mathbf{n}(\check{s})) + \sum_{\check{s}} (\mathbf{m}(\check{s})-\mathbf{n}(\check{s}))(\Psi(\mathbf{m}(\check{s})) - \Psi(\sum_{\check{s}} \mathbf{m}(\check{s})))$.
\Statex \emph{Note 2}: $\Gamma(.)$ and $\Psi(.)$ denote the \emph{gamma} and \emph{digamma} functions, respectively. $[\check{s}=s']$ denotes the Iverson bracket, a number that is $1$ if the condition in square brackets is satisfied, and $0$ otherwise.
\end{algorithmic}
\end{algorithm}

\begin{figure}[!h]
\centering
\includegraphics[height = 0.6\textwidth, width=0.9\textwidth]{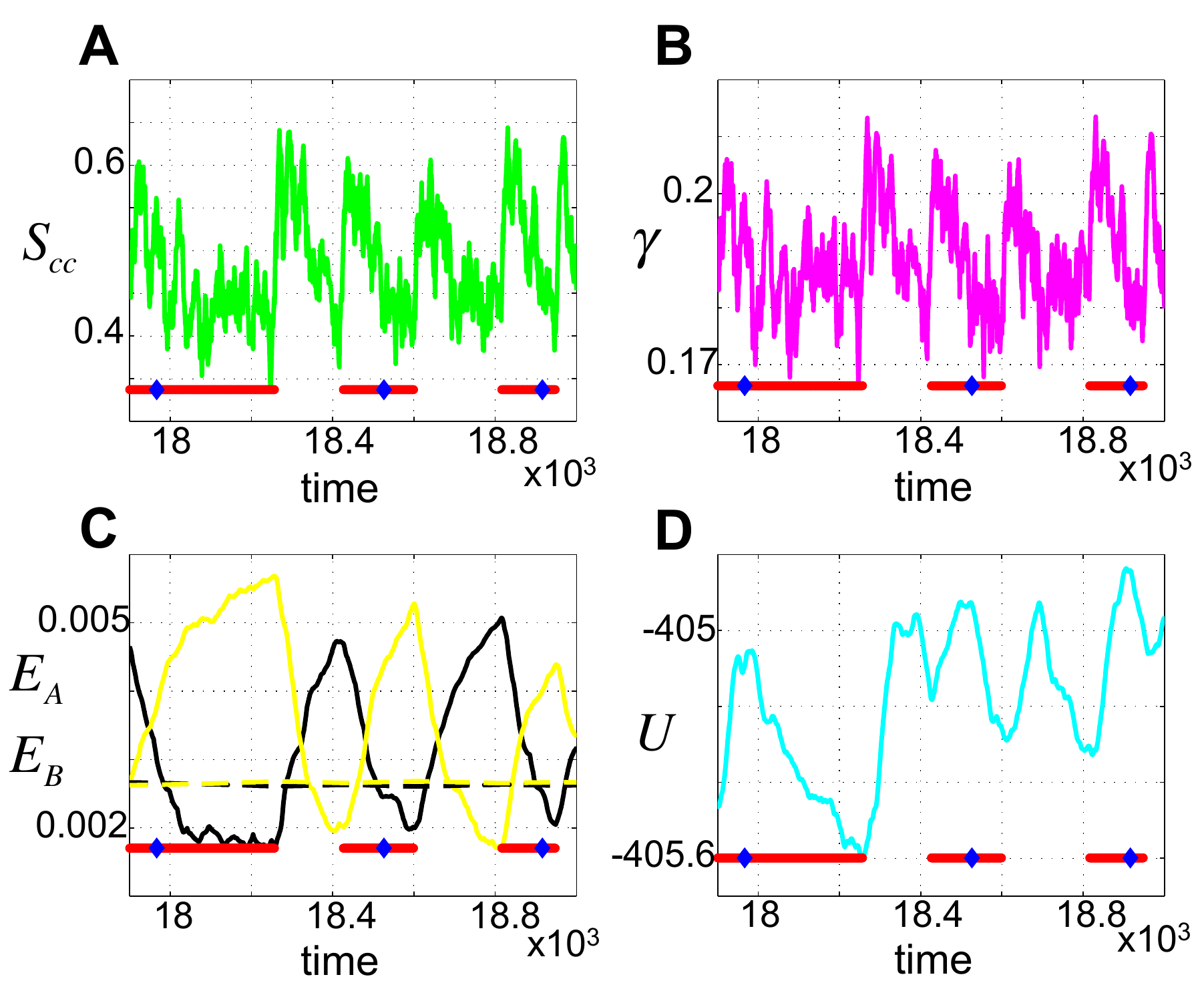}
\caption{{\bf Time-series of relevant signals in the surprise-modulated belief update (Algorithm~\ref{alg:modulated-maze}) applied to the maze exploration task.} All the curves have been smoothed with an exponential moving average (EMA) with a decay constant $0.1$. The plots are shown for $1100$ time steps (horizontal axis) toward the end of a simulation with $20000$ time steps. The agent visits environments $\mathcal{A}$ and $\mathcal{B}$ \emph{equally often} and spends \emph{on average} $200$ time steps in each environment before a switch occurs. Red bars indicate the time that the agent explores environment $\mathcal{A}$. Blue diamonds indicate $100$ time steps after a change point from $\mathcal{B}$ to $\mathcal{A}$. \textbf{A.} Confidence-corrected surprise $S_{cc}$ [Eq~(\ref{eq:AverageShannonModified})] (green) increases at switch points and decreases (with fluctuations) till the next change point. \textbf{B.} The parameter $\gamma$ (magenta) increases with the surprise at the change points and causes the next data samples to be more effective on belief update than the samples before the change point. \textbf{C.} The estimation errors for the transition matrix $\hat{T}$, $E_{\mathcal{A}}[t]=||\hat{T}[t]-T_{\mathcal{A}}||_2=256^{-1} \sum_{s,s'} [\hat{T}[t](s,s')-T_{\mathcal{A}}(s,s')]^2$ (solid black) and $E_{\mathcal{B}}[t]=||\hat{T}[t]-T_{\mathcal{B}}||_2$ (solid yellow) while in environment $\mathcal{A}$ and $\mathcal{B}$, respectively, indicate a rapid adaptation to the new environment after the change points. The dashed black and yellow lines correspond to the estimation errors $E_{\mathcal{A}}$ and $E_{\mathcal{B}}$, respectively, when the naive Bayes rule (as a control experiment) is used for belief update. The naive Bayes rule converges to a stationary solution (no significant change in the estimation error after a switch of environment). \textbf{D.} The model uncertainty (light blue) increases for a few time steps following a change in the environment, an alert that the current model might be wrong. It then starts decreasing as the agent becomes more certain in the new environment.}
\label{fig:Fig 4}
\end{figure}

Similar to the Gaussian mean estimation task, surprise is initially high and slowly decreases as the agent learns the topology of the environment (Fig~\ref{fig:Fig 4}A). When the environment is switched, the sudden increase in the surprise signal (Fig~\ref{fig:Fig 4}A) causes the parameter $\gamma$ to increase (Fig~\ref{fig:Fig 4}B). This is equivalent to discounting previously learned information and results in a quick adaptation to the new environment. To quantify the adaptation to the new environment, we compare the state transition probabilities of the current model with the true transition probabilities of the two environments. We find that the estimation error of the state transition probabilities in the new environment is quickly reduced after the switch points (Fig~\ref{fig:Fig 4}C). Following a change point, the model uncertainty, measured as the entropy of the current belief about the state transition probabilities, increases indicating that the current model of the topology is inaccurate (Fig~\ref{fig:Fig 4}D). A few time steps later the uncertainty slowly decreases, indicating increased confidence in what is learned in the new environment.

\begin{figure}[!h]
\centering
\includegraphics[height = 0.5\textwidth, width=0.8\textwidth]{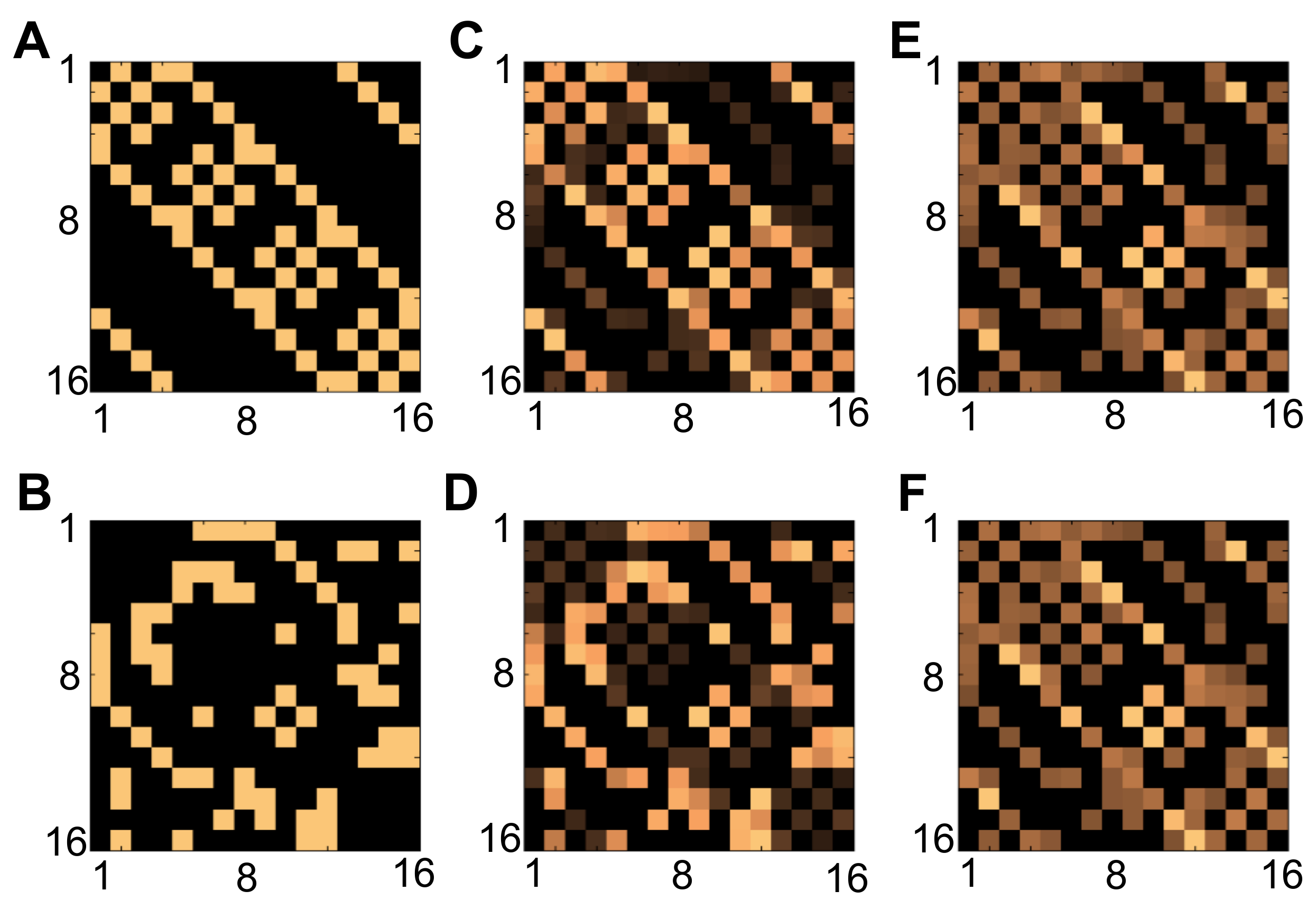}
\caption{{\bf True and estimated state transition probabilities in the maze exploration task.} The color intensity for each entry $(s,s')$ represents the probability of transition from a current state $s$ (row) to a next state $s'$ (column). \textbf{A.} The \emph{true} state transition probability matrix $T_{\mathcal{A}}(s,s')$ in environment $\mathcal{A}$. Each row $T_{\mathcal{A}}(s,:)$ has only four non-zero entries (small squares with the light brown color) whose position indicate the neighboring rooms of state $s$ in environment $\mathcal{A}$. Note that $\sum_{\check{s}} T_{\mathcal{A}}(s,\check{s})=1 \ , \forall s$. \textbf{B.} The true state transition probability matrix $T_{\mathcal{B}}(s,s')$ for the environment $\mathcal{B}$ which has a different topology compared to $\mathcal{A}$. \textbf{C.} The \emph{estimated} state transition probability matrix $\hat{\mathcal{T}}_{\mathcal{A}}$ when the surprise-modulated Algorithm~\ref{alg:modulated-maze} is used for belief update. $\hat{\mathcal{T}}_{\mathcal{A}}=K^{-1}\sum_{k=1}^{K} \hat{T}[t_{\mathcal{B}\rightarrow \mathcal{A}}^k + 100]$ is calculated by averaging the estimated transition matrix $\hat{T}[t]$ at $100$ time steps after each of $K$ change points $t_{\mathcal{B}\rightarrow \mathcal{A}}^k$. Here, $t_{\mathcal{B}\rightarrow \mathcal{A}}^k$ denotes the $k$-th time that the environment is changed from $\mathcal{B}$ to $\mathcal{A}$ and \emph{has remained unchanged} for at least the next $100$ time steps (relevant time points are indicated by blue diamonds in Fig~\ref{fig:Fig 4}). The similarity between $\hat{\mathcal{T}}_{\mathcal{A}}$ and $T_\mathcal{A}$ indicates that Algorithm~\ref{alg:modulated-maze} enables the agent to quickly adapt to environment $\mathcal{A}$ once a switch from $\mathcal{B}$ to $\mathcal{A}$ occurs. \textbf{D.} The estimated transition matrix $\hat{\mathcal{T}}_{\mathcal{B}}$ (similarly defined as $\hat{\mathcal{T}}_{\mathcal{A}}$ but for environment $\mathcal{B}$) when Algorithm~\ref{alg:modulated-maze} is used for belief update. Note its similarity to the true matrix $T_{\mathcal{B}}$. \textbf{E-F.} The estimated state transition probability matrices $\hat{\mathcal{T}}_{\mathcal{A}}$ (top) and $\hat{\mathcal{T}}_{\mathcal{B}}$ (bottom) when the naive Bayesian method (as a control experiment) is used for belief update. A Bayesian agent does not adapt well to the new environment after a switch occurs, because it learns a weighted average of true transition matrices $T_{\mathcal{A}}$ and $T_{\mathcal{B}}$, where the weight is proportional to the fraction of time spent in each environment. Since both environments are visited equally in this experiment, the estimated quantities approach $(T_\mathcal{A}+T_\mathcal{B})/2$.}
\label{fig:Fig 5}
\end{figure}

If we look more closely at the model parameters, we find that the surprise-modulated belief update (Algorithm~\ref{alg:modulated-maze}) enables the agent to adjust the estimated state transition probabilities. In Fig~\ref{fig:Fig 5} we compare the estimated and the true transition probabilities $100$ time steps after a switch. Given that the environment is characterized by $64$ different transitions (in a space of $16\times 15=240$ potential transitions), $100$ time steps allow an agent to explore only a fraction of the potential transitions. Nevertheless, $100$ time steps after a switch, the matrix of transition probabilities already resembles that of the present environment (Figs~\ref{fig:Fig 5}C and \ref{fig:Fig 5}D).

The surprise-modulated belief update is a method of quick learning. How well does our SMiLe update rule perform relative to other existing models? We compared it with two well-known models. First, we compared to a naive Bayesian learner which tries to estimate the $240$ state transition probabilities using Bayes rule. Note that, by construction, the naive Bayesian learner is not aware of the switches between the environments. Second, we compared to a hierarchical statistical model that reflects the architecture of the \emph{true world} as in Fig~\ref{fig:Fig 3}. The task is to estimate the $2\times 240$ state transitions in the two environments as well as transition probabilities between the environments $p_{\mathcal{A}\rightarrow \mathcal{B}}$ and $p_{\mathcal{B}\rightarrow \mathcal{A}}$ by an online EM algorithm.

For the naive Bayesian learner, we find that its behavior indicates a steady increase in certainty, regardless of how surprising the samples are. In other words, it is incapable of changing its belief after it has sufficiently explored the environments (Fig~\ref{fig:Fig 4}C). The state transition probabilities are estimated by averaging over the true parameters of both environments, where the weight of averaging is determined by the fraction of time spent in the corresponding environment (Figs~\ref{fig:Fig 5}E and \ref{fig:Fig 5}F). 

\begin{figure}[!h]
\centering
\includegraphics[width=1\textwidth, height=0.4 \textwidth]{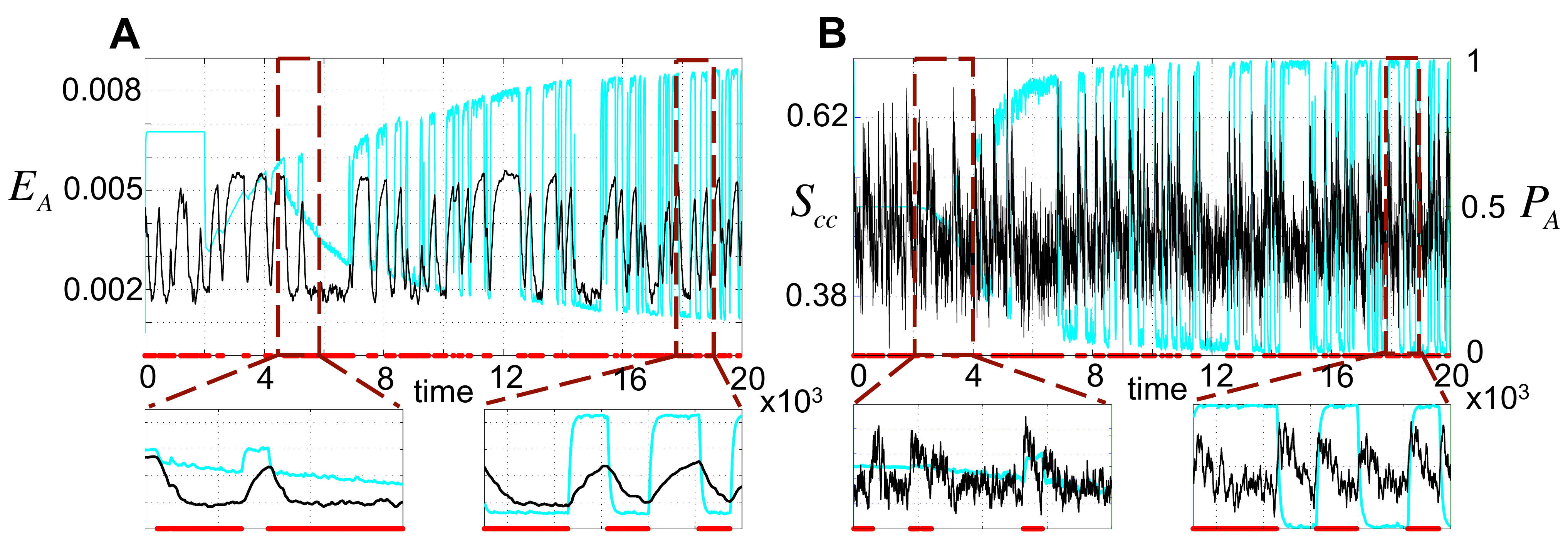}
\caption{{\bf Comparison of surprise-modulated belief update with an online EM algorithm for the hierarchical Bayesian model.} \textbf{A.} The estimation error $E_\mathcal{A}$ (vertical axis) of state transition probabilities within environment $\mathcal{A}$ versus time (horizontal axis), for surprise-modulated belief update (black) and online EM learner (blue). Bottom plots depict zooms during the early (left) and late (right) phases of a simulation of $20000$ time steps. In the early phase of learning (bottom left), the surprise-modulated belief update enables the agent to quickly learn model parameters after a switch to environment $\mathcal{A}$ (indicated by red bars). In the late phase of learning (right), however, the online EM algorithm adapts to the new environment faster and more accurately than the surprise-modulated belief update. \textbf{B.} The inferred probability $P_\mathcal{A}$ of being in environment $\mathcal{A}$ (blue, right vertical axis) used in the online EM algorithm, and the confidence-corrected surprise $S_{cc}$ (black, left vertical axis) used in the surprise-modulated belief update.}
\label{fig:Fig 6}
\end{figure}

The comparison of our surprise-modulated belief update with the online EM algorithm~\citep{mongillo2008online} for the hierarchical Bayesian model associated with the changing environments provides several insights (see Fig~\ref{fig:Fig 6}). First, already after less than $1000$ time steps, the estimation error for environment $\mathcal{A}$ during short episodes in environment $\mathcal{A}$ drops below $E_\mathcal{A}=0.002$. The online EM algorithm takes 10 times longer to achieve the same level of accuracy. While the solution of the SMiLe rule in the long run is not as good as that of the online EM algorithm, our algorithm benefits from a reduced computational complexity and simpler implementation.  

To further investigate the ability of an agent to adapt to the new environment after a switch, we analyzed performance as a function of two free parameters that control the setting of the task: (i) the fraction of time spent in environment $\mathcal{A}$, and (ii) the average time spent in environment $\mathcal{A}$ before a switch to $\mathcal{B}$ occurs. To do so, we calculate the average estimation error in state transition probabilities $64$ time steps after a switch occurs. We consider only those switches after which the agent stays in that environment for \emph{at least} $64$ time steps. Note that $64$ is the minimum number of time steps that is required to ensure that all possible transitions from $16$ room to their $4$ neighbors \emph{could} occur. A smaller estimation error for a given pair of free parameters indicates a faster adaptation to the new environment for that setting. 

\begin{figure}[!h]
\centering
\includegraphics[width=1\textwidth, height=0.3\textwidth]{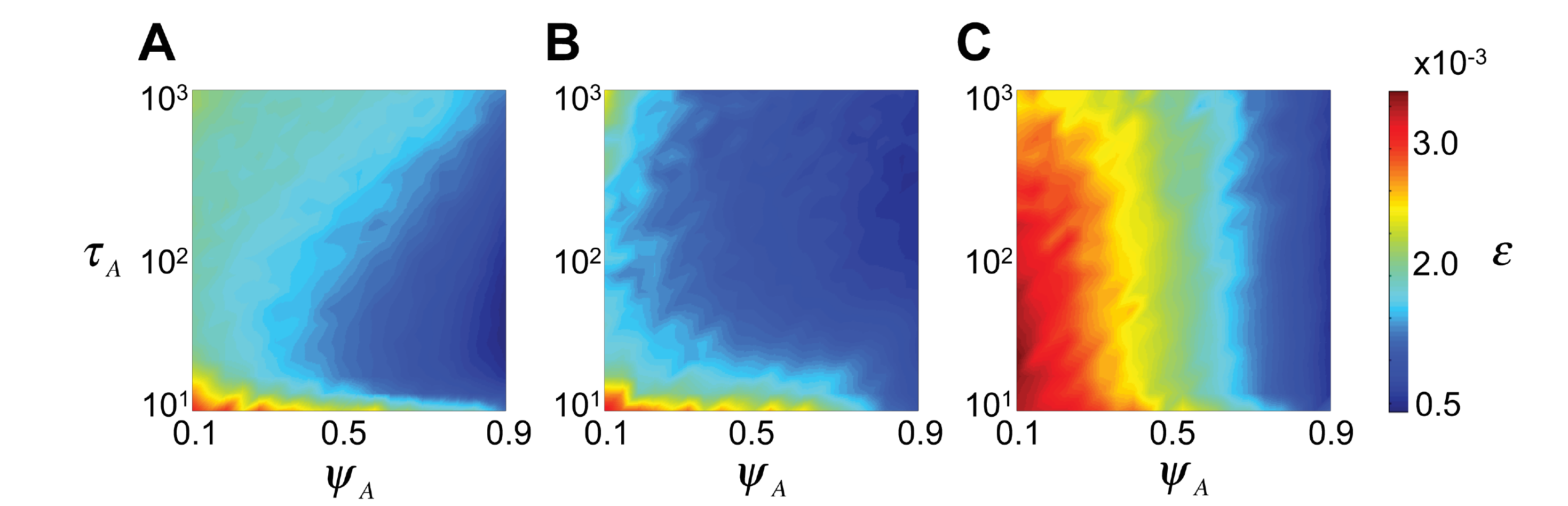}
\caption{{\bf The estimation error $\epsilon$ in the maze exploration task,} as a function of (1) the average time spent in environment $\mathcal{A}$ before a switch to environment $\mathcal{B}$ ($\tau_\mathcal{A}=\Delta t / p_{\mathcal{A}\rightarrow\mathcal{B}}$, vertical axis) and (2) the fraction of time spent in environment $\mathcal{A}$ ($\psi_\mathcal{A} = P_{\mathcal{B}\rightarrow \mathcal{A}} /(P_{\mathcal{B}\rightarrow \mathcal{A}} + P_{\mathcal{A}\rightarrow \mathcal{B}})$, horizontal axis). \textbf{A.} The average estimation error (of state transition probabilities), $64$ time steps after a switch from $\mathcal{B}$ to $\mathcal{A}$, when surprise-modulated belief update (Algorithm \ref{alg:modulated-maze}) is used for learning. The spread of blue color (lower estimation error) illustrates that the surprise-modulated belief update enables an agent to quickly adapt to the environment visited after a switch. For each pair $(\tau_\mathcal{A}, \psi_\mathcal{A})$, the simulation is repeated for $20$ episodes, each consisting of $20000$ time steps. In each episode a different rearrangement of rooms for building environment $\mathcal{B}$ is used to make sure that the result is not biased by a specific choice of this environment. \textbf{B.} The average estimation error when the online EM algorithm is used for learning the hierarchical statistical model. \textbf{C.} The average estimation error when the naive Bayesian learner is used for belief update. The estimation error for this model is mainly determined by the fraction of time spent in environment $\mathcal{A}$ (i.e., $\psi_\mathcal{A}$). The estimation error decreases with the time spent in environment $\mathcal{A}$, regardless of the time scale of stability determined by $\tau_\mathcal{A}$.}
\label{fig:Fig 7}
\end{figure}

We found that the surprise-modulated belief enables an agent to quickly readjust its estimation of model parameters, even if the fraction of time spent in an environment is relatively short. In that sense, it behaves similarly to the approximate hierarchical Bayesian approach (online EM algorithm). This is not, however, the case for a naive Bayesian learner whose estimation error in each environment depends on the fraction of time spent in the corresponding environment (see Fig~\ref{fig:Fig 7}). 

The naive Bayesian learner suffers from low accuracy in estimation and cannot adapt to environmental changes. A full hierarchical Bayesian model, however, requires prior information about the task and is computationally demanding. For example, the computational load of the hierarchical Bayesian model increases with the number $N$ of environments between which switching occurs. The surprise-modulated belief update, however, balances accuracy and computational complexity: computational complexity remains, by construction, independent of the number of switched environments. In other words, since we accept from the beginning that our model of the world will be approximate and structurally incomplete, the model can perform reasonably well after having seen a small number of data samples.

\section{Discussion}

Surprise is a widely used concept describing a range of phenomena from unexpected events to behavioral responses. Existing approaches to quantifying surprise are either data-centric~\citep{shannon1948mathematical, tribus1961information, palm2012novelty} or model-centric~\citep{baldi2010bits, itti2005bayesian}, and may be objective in a known model of the world~\citep{shannon1948mathematical, tribus1961information} or subject-dependent and rely on a learned model of the world~\citep{palm2012novelty, baldi2010bits, itti2005bayesian}, but are always linked to uncertainty. We emphasize that in order for surprise to be behaviorally meaningful, it has to be defined for a \emph{single} data sample such that an organism can respond to a single event. In contrast, information theoretic quantities, such as data entropy and mutual information, are usually defined as average quantities.   

Based on our definition of surprise, we proposed a new framework for surprise-driven learning. There are two components to this framework: (i) a confidence-adjusted surprise measure to capture environmental statistics as well as the commitment of the subject to his belief, and (ii) the surprise-minimization learning rule, or SMiLe-rule, which dynamically adjusts the balance between new and old information without prior assumptions about the temporal statistics in the environment. Within this framework, surprise is a single subject-specific variable that determines a subject's propensity to modify existing beliefs. This algorithm is suitable for learning in complex environments that are either stable or undergo gradual or sudden changes using a world model that may not match the complexity of the world. Sudden changes are signalled by high surprise and result in placing more weight on new information. The significance of the proposed method is that it neither requires knowledge of the full Bayesian model of the environment nor any prior assumption about the temporal statistics in the environment. Moreover, it provides a simple framework that could potentially be implemented in a neurally plausible way using probabilistic population codes~\citep{ma2006bayesian, beck2008probabilistic}.

\subsection{Relation to Bayesian surprise}

One of the existing approaches for measuring surprise is \emph{Bayesian surprise}~\citep{baldi2010bits, itti2009bayesian} which is generally defined as a distance or dissimilarity measure between prior and posterior beliefs where the updating is performed according to Bayes' rule.  With this measure, a data sample $X$ is more surprising than a data sample $X'$ if it causes a larger change in the subject's belief. One of the shortcomings of the Bayesian surprise is that it can be evaluated only \emph{after} the learning step (i.e., once we have changed our belief from prior to posterior). However, behavioral and neural responses indicate that surprise is nearly concurrent with the unexpected event, since physiological signals such as the P300 component of the EEG occur within less than 400 ms.
Our working hypothesis is that the brain evaluates surprise even before recognition, inference or learning occurs. We thus need to evaluate surprise \emph{before} we update our belief so that surprise may control learning rather than emerge from it. This property is fulfilled in the confidence-corrected surprise measure introduced in this paper.

Information content and Bayesian surprise are two distinct yet complementary approaches to measuring surprise. Information content is about \emph{data} as it captures the inherent likelihood of a piece of data given a model. While its evaluation is rapid given a world model, we normally (outside an engineered lab environment) do not have access to the true underlying parameter $\theta^*$. Bayesian surprise is about a \emph{model}. Since it measures the change in belief (i.e., in the model parameters), it is subject-dependent and does not require knowing $\theta^*$. However, it is computed only after learning, whereas neural data suggests response to surprise within 400 ms. Our definition of confidence-corrected surprise combines these two measures to use their complementary benefits and overcome their shortcomings (see {\bf Mathematical Methods}).

\subsection{New versus old information}

The proposed algorithm's performance is primarily driven by two features: (i) the algorithm adaptively increases the influence of new data on the belief update as a function of how surprising the data was; and (ii) the algorithm increases model uncertainty in the face of surprising data thus increasing the influence of new data on current \emph{and} future belief updates. The importance of the first point has been recognized and incorporated previously~\citep{nassar2012rational, pearce1980model}. The second point is particularly worth noting: a surprising sample not only signals a potential change, it also signals that our current model may be wrong, so that we should be \emph{less} certain about its accuracy. This increase in model uncertainty implies discounting the influence of past information in current and future belief updates.

Both humans and animals adaptively adjust the relative contribution of old and newly acquired data on learning~\citep{behrens2007learning, nassar2012rational, krugel2009genetic, pearce1980model} and rapidly adapt to changing environments~\citep{pearce1980model, wilson1992restoration, holland1997brain}. Standard Bayesian and reinforcement learning models in humans~\citep{tenenbaum2001structure} or animals~\citep{dayan2000learning, kakade2002acquisition} assume a stable environment and are slow to adapt to sudden changes in the environment. To quickly learn in dynamic environments, models need to include a way to detect and respond to sudden changes.

A full (hierarchical) Bayesian approach works only if the subject is aware of the correct model of the task, (e.g., the time scale of change in the environment or the number of environments between which switches occur). Calculating the probability of a change point in a Gaussian estimation task~\citep{nassar2010approximately}, estimating the volatility of the environment in a reversal learning task~\citep{behrens2007learning}, and dynamically forgetting the past information with a controlled time constant~\citep{ruter2012paradoxical} are all examples of addressing learning in changing environments without explicit knowledge of the full Bayesian model.

In changing environments, hierarchical Bayesian models outperform the standard delta-rule with a fixed learning rate. However, hierarchical models either make assumptions about how fast the world is changing on average or about the underlying data generating process, in order to accurately detect a change in the environment. While our proposed surprise-based algorithm may not be theoretically optimal, it approximates the optimal (hierarchical) Bayesian solution without making any such assumption.

\subsection{Model uncertainty}

The ability of our proposed method to increase model uncertainty solves a common problem in standard Bayesian learning, namely, a model uncertainty or a learning rate approaching zero when the number of data samples increases. This is particularly prominent in Bayes' rule which is derived under the assumption of \emph{stationarity} and which thus reduces  uncertainty in each step no matter how surprising a sample is. The SMiLe rule [Eq~(\ref{eq:ModifiedBayesian2})] guarantees that a small model uncertainty remains even after a long stationary period. This remaining uncertainty ensures that an organism can still detect a change even after having spent an extensive amount of time in a given environment (see Fig~\ref{fig:Fig 4}C). One might argue, that reducing the learning rate to zero after extensive training is desirable under certain conditions as it corresponds to the well-documented phenomenon of overtraining whereby an organism no longer responds to changes in goal value. We would argue that this insensitivity is a consequence of behavioral control being handed over to the habitual system and thus to a different neural substrate~\citep{balleine2010human, balleine1998goal, redgrave2010goal}.

\subsection{Potential applications}

Surprise minimization is a more general approach to learning than learning by reward prediction error. Recent approaches in reward learning suggest using a scaled reward prediction error~\citep{preuschoff2007adding}. A recurring problem in reward-based learning is the observation that subjects use different learning rates on a trial-by-trial basis even in stable environments. Researchers typically assume an average learning rate for fitting data. Note that in our approach, the learning rate varies naturally as a function of the last data point (as it should) while keeping the subject-specific parameter $m$ constant.

Note that both confidence-corrected surprise and the SMiLe rule have wide-reaching implications outside the framework presented here. On the one hand, our surprise measure can not only \emph{modulate} learning, but can be used as a \emph{trigger} signal for an algorithm that needs to choose between several uncertain states or actions as is the case in change point detection~\citep{nassar2010approximately, wilson2013mixture, ruter2012paradoxical}, memory and cluster formation~\citep{gershman2015novelty}, exploration/exploitation tradeoff~\citep{cohen2007should, jepma2011pupil}, novelty detection~\citep{knight1996contribution, bishop1994novelty}, and network reset~\citep{bouret2005network}. On the other hand, the SMiLe-rule could add flexibility in learning and replace existing learning algorithms in scenarios where dynamically balancing old and new information is desired. This includes fitting $\gamma$ to behavioral data without computing surprise or controlling $\gamma$ by something other than surprise. Replacing the full Bayesian model of a learning task in changing environment with the SMile rule simplifies calculations, which makes the SMiLe-framework suitable for fitting relevant parameters to behavioral data. 

\subsection{Experimental predictions}

There is ample evidence for a neural substrate of surprise. Existing measures of expectation violations such as absolute and variance-scaled reward prediction errors~\citep{schultz2016dopamine, schultz2015neuronal}, unexpected uncertainty~\citep{yu2005uncertainty} and risk prediction errors~\citep{preuschoff2008human} have been linked to different neuromodulatory systems. Among those, the \emph{noradrenergic system} has emerged as a prime candidate for signaling unexpected uncertainty and surprise: noradrenergic neurons respond to unexpected changes such as the presence of a novel stimulus, unexpected pairing of stimulus with a reinforcement during conditioning, and reversal of the contingencies~\citep{sara1991plasticity, sara1994locus, vankov1995response, aston1997conditioned}. The P300 component of the event-related potential~\citep{pineda1997human, missonnier1999automatic} which is associated with novelty~\citep{donchin1978cognitive} and surprise~\citep{jaskowski1994suspense} is modulated by Noradrenaline. It also modulates pupil size~\citep{costa2016more} as a physiological response to surprise. The dynamics of the noradrenergic system are fast enough to quickly respond to unexpected events~\citep{rajkowski1994locus, clayton2004phasic, bouret2004reward}, a functional requirement for surprise to control learning; see~\citep{sara2009locus, bouret2005network, aston2005integrative} for a review. We predict that in experiments with changing environments, the activity of Noradrenaline should exhibit a high correlation with the confidence-corrected surprise signal. 

Note that Acetylcholine (ACh), on the other hand, is a candidate neuromodulator for encoding expected uncertainty~\citep{yu2005uncertainty} and thus is linked to the model uncertainty (although it might also be linked to other forms of uncertainty such as environmental stochasticity). 

A variety of experimental findings are consistent with and can be explained by our definition of confidence-corrected surprise and the SMiLe rule. It has been shown both theoretically~\citep{yu2005uncertainty} and empirically~\citep{gu2002neuromodulatory} that Noradrenaline and ACh interact such that ACh sets a threshold for Noradrenaline to indicate fundamental changes in the environment~\citep{yu2005uncertainty}. This is consistent with our hypothesis that if an agent is uncertain about its current model of the world, unexpected events are perceived as less surprising than when the agent is almost certain about the model (which is the key idea behind the confidence-corrected surprise). The impairment of adaptation to contextual changes due to Noradrenaline depletion~\citep{sara1998learning} can be explained by the incapability of subjects to respond to surprising events signaled by Noradrenaline. The absence/suppression of ACh (low model uncertainty) implies little or no variability of the environment so that even small prediction error signals are perceived as surprising~\citep{jones1995effect}, consistent with the excessive activation of noradrenergic system in such situations. 

Moreover, there is empirical evidence that Noradrenaline and ACh both affect synaptic plasticity in the cortex and the hippocampus~\citep{gu2002neuromodulatory, bear1986modulation}, suppress cortical processing~\citep{kimura1999acetylcholine, kobayashi2000selective}, and facilitate information processing from thalamus to the cerebral cortex~\citep{gil1997differential, hasselmo1996encoding, hsieh2000differential}. This is consistent with our theory that surprise balances the influence of newly acquired data (thalamocortical pathway) and old information (corticocortical pathway) during belief update.

In summary, we proposed a measure of surprise and a surprise-modulated belief update algorithm that can be used for modeling how humans and animals learn in changing environments. Our results suggest that the proposed method can approximate an optimal hierarchical Bayesian learner, with significantly reduced computational complexity, but at the cost of an imperfect model of the world. Our model provides a framework for future studies on learning with surprise. These include computational studies, such as how the proposed model can be neurally implemented, neurobiological studies, such as unraveling the interaction between different neural circuits that are functionally involved in learning under surprise, and behavioral studies with human subjects.

\section{Mathematical Methods}

In this section we provide mathematical explanations or proofs for statements made in the `Results' Section.

\subsection{The scaled likelihood is the posterior belief under a flat prior}

Assume that all model parameters $\theta$ stay in some bounded convex interval of volume $A$.
The volume $A$ can be arbitrarily large.
Given a data sample $X$, the posterior belief $p^{flat}(\theta|X)$ about the model parameters $\theta$ (derived by the Bayes rule) under the assumption of a flat prior $\hat{\pi}_0(\theta)=1/A$ is:
\begin{equation}
p^{flat}(\theta|X)= \frac{p(X|\theta)\hat{\pi}_0(\theta)}{\int_\theta p(X|\theta)\hat{\pi}_0(\theta) \ d\theta} = \frac{p(X|\theta)}{\int_\theta p(X|\theta) \ d\theta} = \frac{p(X|\theta)}{||p_X||}=\hat{p}_X(\theta),
\end{equation}
where $||p_X||=\int_\theta p(X|\theta) d\theta$ is a data-dependent constant. Therefore, the scaled likelihood $\hat{p}_X(\theta)$ is the posterior under a flat prior. Note that the result is independent of the volume of $A$ so that we take the limit of $A\to \infty$.


\subsection{Confidence-corrected surprise increases with Shannon surprise and Bayesian surprise}

The confidence-corrected surprise in Eq.~(\ref{eq:AverageShannonModified}) can be expressed as: 
\begin{equation}
S_{cc}(X;\pi_n) = -\int_\theta \pi_n(\theta) \ln p(X|\theta) + \ln ||p_X|| - H(\pi_n),
\label{eq:ConfSurpExpanded}
\end{equation}
where $||p_X||$ is a data-dependent constant defined above, and $H(\pi_n)= -\int_\theta \pi_n(\theta) \ln \pi_n(\theta) d\theta$ denotes the entropy of the current belief.
Let us  call the first term $-\int_\theta \pi_n(\theta) \ln p(X|\theta)$ in Eq.~(\ref{eq:ConfSurpExpanded}) the \emph{raw} surprise $S_{raw}(X;\pi_n)$ of a data sample $X$:
\begin{equation}
S_{raw}(X;\pi_n) = - \int_\theta \pi_n(\theta) \ \ln p(X|\theta)  \ d\theta,
\label{eq:AverageShannon}
\end{equation}
We want to show that the raw surprise $S_{raw}(X;\pi_n)$ in Eq.~(\ref{eq:AverageShannon}) increases with the Shannon surprise and the Bayesian surprise.

The {\bf Bayesian} surprise~\citep{itti2009bayesian, baldi2010bits} measures a  change in belief
induced by the observation of a new data sample $X=X_{n+1}$. It is defined as a KL divergence $D_{KL}[\pi_n||\pi_{n+1}]$ between the prior belief $\pi_n$ and the posterior belief $\pi_{n+1}(\theta)$ that is calculated from the naive Bayes rule
\begin{equation}
\pi_{n+1}(\theta) =\frac{p(X|\theta)\pi_n(\theta)}{\int_\theta p(X|\theta)\pi_n(\theta) \ d \theta}.
\label{eq:BayesRule}
\end{equation}
The {\bf Shannon} surprise~\citep{shannon1948mathematical, tribus1961information, palm2012novelty}
is the the information content of data point $X=X_{n+1}$  calculated with the current world model
of the subject.
If the true model of the world (i.e., $\theta^*$) is known, the \emph{information content} $-\ln p(X|\theta^*)$ for a specific outcome $X\in\mathcal{X}$ is the negative log-likelihood of  this data point~\citep{tribus1961information, shannon1948mathematical, palm2012novelty}. In other words,
the occurrence of a rare (i.e., unlikely to occur) data sample $X$ is surprising.
As the information content relates to the \emph{true} probabilities $p(X|\theta^*)$ of samples in the \emph{real} world, it is an \emph{objective}, model-independent, measure of unexpectedness.
However, we work under the assumption that
the true set of parameters $\theta^*$, and thus the true probability $p(X|\theta^*)$, is not known
to the observer, such that it is difficult to evaluate the exact information content of a data sample $X$. The Shannon surprise is defined as the negative-log-marginal-likelihood $-\ln Z(X)$, where $Z(X)=\int_\theta p(X|\theta)\pi_n(\theta) d\theta$ is the probability of data sample $X$ after marginalizing out all the possible model parameters.

In the following, small numbers above an equality sign refer to equations in the text. The raw surprise $S_{raw}(X;\pi_n)$ in Eq~(\ref{eq:AverageShannon})
is a linear combination of both Bayesian surprise and Shannon surprise, because
\begin{eqnarray}
S_{raw}(X;\pi_n) &\overset{(\ref{eq:AverageShannon})}{=}& -\int_\theta \pi_n(\theta) \ln p(X|\theta) \ d\theta \nonumber \\
&\overset{(\ref{eq:BayesRule})}{=}& -\int_\theta \pi_n (\theta) \ln \big[ \frac{\pi_{n+1} (\theta) \left( \int_\theta p(X|\theta) \pi_{n}(\theta) \ d\theta \right) }{\pi_{n}(\theta)} \big] \ d\theta \nonumber \\ &=& D_{KL}[\pi_n||\pi_{n+1}] -\ln \big[ \int_\theta p(X|\theta) \pi_{n}(\theta) \ d\theta \big],
\label{eq:rawVersusBayes}
\end{eqnarray}
where the first term $D_{KL}[\pi_n||\pi_{n+1}]$ stands for the Bayesian surprise, and the second term $-\ln \big[ \int_\theta p(X|\theta) \pi_{n}(\theta) \ d\theta \big]$ stands for the Shannon surprise.
Therefore, the raw surprise $S_{raw}(X;\pi_n)$ in Eq~(\ref{eq:rawVersusBayes}) combines both the data-driven approach of Shannon (information content) and the model-change driven approach for Itti and Baldi (Bayesian surprise). 

\subsection{Less likely data lead to a larger surprise $S_{cc}$}

Our proposed confidence-corrected surprise measure $S_{cc}(X;\pi_n)$ in Eq~(\ref{eq:AverageShannonModified}) inherits the property of the raw surprise $S_{raw}(X;\pi_n)$ in Eq~(\ref{eq:rawVersusBayes})
which in turn is a linear combination of Bayesian surprise and Shannon surprise.
Therefore it inherits the properties of the Shannon surprise.
In particular, for a fixed opinion $\pi_n$, a less likely data point leads to a larger surprise.

\subsection{Committed subjects are more surprised than uncommitted ones}

The value of the confidence-corrected surprise Eq.~(\ref{eq:ConfSurpExpanded}) depends on a subject's
commitment to  her belief.
The commitment to  the current model of the world is represented by the negative \emph{entropy} $-H(\pi_n)=\int_\theta \pi_n(\theta) \ln \pi_n(\theta) d\theta$.
Eq.~(\ref{eq:ConfSurpExpanded}) shows that 
the confidence-corrected surprise 
decreases with  entropy indicating an  increases with commitment.
Therefore, given the same likelihood of the data under two different world models,
the subject with a stronger commitment (smaller entropy) is more surprised than the subject with a weaker commitment (higher entropy); cf. the example of Fig. \ref{fig:Fig 1}A.

Intuitively, if we are uncertain about what to expect (because we have not yet learned the structure of the world), receiving a data sample that occurs with low probability under the present model is less surprising than a low-probability sample in a situation when we are almost certain about the world (see Fig~\ref{fig:Fig 1}A).

\subsection{Calculation of surprise for the example of CEO election}

If candidate 1 is selected, the surprise of colleague B, $S_{cc}(X=1;\pi^B)$ is bigger by $0.75 \ln \frac{(1-\epsilon)}{\epsilon / 3} > 0$ than the surprise of colleague A, $S_{cc}(X=1;\pi^A)$. Both colleagues are equally committed to their believes, but the outcome ``candidate 1'' is less likely for colleague B than A. The evaluation of surprise yields
\begin{adjustwidth}{-1cm}{} 
\begin{eqnarray}
S_{cc}(X=1;\pi^B) - S_{cc}(X=1;\pi^A) &=& \sum_k \pi^B(\theta_k) \ln \frac{\pi^B(\theta_k)}{\hat{p}_{X=1}(\theta_k)} - \sum_k \pi^A(\theta_k) \ln \frac{\pi^A(\theta_k)}{\hat{p}_{X=1}(\theta_k)} \nonumber\\
&=& -H(\pi^B) + H(\pi^A) + \sum_k \left( \pi^A(\theta_k) - \pi^B(\theta_k) \right) \ln \hat{p}_{X=1}(\theta_k) \nonumber \\
&=& \sum_k \left( \pi^A(\theta_k) - \pi^B(\theta_k) \right) \ln \hat{p}_{X=1}(\theta_k) \nonumber \\
&=& 0.75 \ln \frac{(1-\epsilon)}{\epsilon / 3} > 0. 
\end{eqnarray}
\end{adjustwidth}
Therefore B is more surprised than A.
 
For colleague A, surprise of the outcome ``candidate 2'', $S_{cc}(X=2;\pi^A)$, is bigger by $0.5 \ln \frac{1-\epsilon}{\epsilon / 3} > 0$ than the surprise of outcome ``candidate 1'' (his favorite), $S_{cc}(X=1;\pi^A)$ because the second candidate is less likely in his opinion; cf. point (ii) at the beginning of the subsection ``Definition of Surprise'': 
\begin{eqnarray}
S_{cc}(X=2;\pi^A) - S_{cc}(X=1;\pi^A) &=& \sum_k \pi^A(\theta_k) \left( \ln \frac{\pi^A(\theta_k)}{\hat{p}_{X=2}(\theta_k)} - \ln \frac{\pi^A(\theta_k)}{\hat{p}_{X=1}(\theta_k)} \right) \nonumber\\
&=& \sum_k \pi^A(\theta_k) \ln \frac{\hat{p}_{X=1}(\theta_k)}{\hat{p}_{X=2}(\theta_k)} \nonumber \\
&=& 0.75 \ln \frac{1-\epsilon}{\epsilon / 3} + 0.25 \ln \frac{\epsilon / 3}{1-\epsilon} \nonumber \\
&=& 0.5 \ln \frac{1-\epsilon}{\epsilon / 3} > 0. 
\end{eqnarray}

More importantly, however, if the second candidate wins, the surprise of colleague A, $S_{cc}(X=2;\pi^A)$, is bigger than that of colleague C, $S_{cc}(X=2;\pi^C)$, even though both have assigned the same low likelihood to the second candidate. The evaluation of surprise yields 

\begin{adjustwidth}{-1cm}{} 
\begin{eqnarray}
S_{cc}(X=2;\pi^A) - S_{cc}(X=2;\pi^C) &=& \sum_k \pi^A(\theta_k) \ln \frac{\pi^A(\theta_k)}{\hat{p}_{X=2}(\theta_k)} - \sum_k \pi^C(\theta_k) \ln \frac{\pi^C(\theta_k)}{\hat{p}_{X=2}(\theta_k)} \nonumber\\
&=& \sum_k \left( \pi^C(\theta_k) - \pi^A(\theta_k) \right)  \ln \hat{p}_{X=2}(\theta_k) \nonumber \\
 && -H(\pi^A) + H(\pi^C).
\label{eq:calcul}
\end{eqnarray}
\end{adjustwidth}
The terms with the $\ln$ in Eq.~(\ref{eq:calcul}) add up to zero (i.e., $- 0.5 \ln \epsilon/3 + 0 \ln (1-\epsilon) + 0.25 \ln \epsilon /3 + 0.25 \ln \epsilon / 3 = 0$), so that we just need to evaluate the entropies to find 
\begin{adjustwidth}{-1cm}{} 
\begin{eqnarray}
S_{cc}(X=2;\pi^A) - S_{cc}(X=2;\pi^C) &=& -H(\pi^A) + H(\pi^C) \nonumber \\
&=& 0.75 \ln 0.75 + 0.25\ln 0.25 + \ln 4 \nonumber \\
&=& 0.75 \ln 3 >0 
\end{eqnarray}
\end{adjustwidth}
In other words, since colleague A is more committed to his opinion than colleague C, i.e. $H(\pi^C) > H(\pi^A)$, he will be more surprised; cf. point (iii):


\subsection{Derivation of the SMiLe rule}

We note that the KL divergence $D_{KL}[a||b]$ is convex with respect to the first argument $a$. Therefore, both the objective function $S_{cc}(X;q)$ in Eq~(\ref{eq:AverageShannonModified}) and the constraint $D_{KL}[q||\pi_n] \leq B$ in the optimization problem in Eq~(\ref{eq:MinSurp}) are convex with respect to $q$, which ensures the existence of the optimal solution.
 
We solve the constraint optimization by introducing a non-negative Lagrange multiplier $\lambda^{-1} \geq 0$ and a Lagrangian 
\begin{eqnarray}
\mathbb{L}(q,\lambda) &=& S_{cc}(X;q) - \frac{1}{\lambda} ( B- D_{KL}[q|| \pi_n] ) \nonumber \\
&\overset{(\ref{eq:ConfSurpExpanded})}{=}& \left< -\ln p(X|\theta) + \ln q(\theta) + \frac{1}{\lambda} \ln \frac{q(\theta)}{\pi_n(\theta)} \right>_q -\frac{B}{\lambda} + \ln ||p||,
\label{eq:Lagrangian}
\end{eqnarray}
where $\left<.\right>_q$ denotes the average with respect to $q$. Similar to the standard approach that is used in support vector machines \citep{scholkopf2002learning}, the Lagrangian $\mathbb{L}$ defined in Eq~(\ref{eq:Lagrangian}) must be minimized with respect to the primal variable $q$ and maximized with respect to the dual variable $\lambda$ (i.e., a saddle point must be found). Therefore the constraint problem in Eq~(\ref{eq:MinSurp}) can be expressed as 
\begin{equation}
\arg \min_{q} \max_{\lambda \geq 0} \ \mathbb{L}(q,\lambda).
\label{eq:optimizationSVM}
\end{equation}

By taking the derivative of $\mathbb{L}$ with respect to $q$ and setting it equal to zero, 
\begin{equation}
\frac{\partial \mathbb{L}}{\partial q} = -\ln p(X|\theta) + \big[ 1+\ln q(\theta) \big] + \frac{1}{\lambda} \big[ 1+\ln \frac{q(\theta)}{\pi_n(\theta)} \big] = 0,
\label{eq:derivative2}
\end{equation}
we find that the Lagrangian in Eq~(\ref{eq:Lagrangian}) is minimized by the SMiLe rule [Eq~(\ref{eq:ModifiedBayesian2})], i.e., $q(\theta) \propto p(X|\theta)^\gamma \pi_n(\theta)^{1-\gamma}$, where $\gamma$ is determined by the Lagrange multiplier $\lambda$: 

\begin{equation}
0 \leq \gamma=\frac{\lambda}{\lambda+1} \leq 1, 
\label{eq:GammaFromLambda}
\end{equation}

Note that the constant $Z(X;\gamma)$ in Eq~(\ref{eq:ModifiedBayesian2}) follows from normalization of $q(\theta)$ to integral one.

\subsection{A larger bound $B>B'$ on belief change implies a bigger $\gamma>\gamma'$ in the SMiLe rule}

For $0 < B < B_{max}$ the solution of optimization problem in Eq~(\ref{eq:MinSurp}) is the
updated belief $q_\gamma$ [Eq~(\ref{eq:ModifiedBayesian2})] with $0<\gamma <1$ satisfying $D_{KL}[q_\gamma||\pi_n]= B$. In order to prove that $B>B'$ implies $\gamma>\gamma'$, we just need to show that $D_{KL}[q_\gamma||\pi_n]$ is an increasing function of $\gamma$ and thus its first derivative with respect to $\gamma$ is always non-negative. 

For this purpose, we first need to evaluate the derivative of $q_\gamma(\theta)$, [Eq~(\ref{eq:ModifiedBayesian2})], with respect to $\gamma$:
\begin{adjustwidth}{-2cm}{}
\begin{eqnarray}
\frac{\partial}{\partial \gamma} q_\gamma(\theta) &=& \frac{1}{Z(X;\gamma)} \ \frac{\partial}{\partial \gamma}  \big[ p(X|\theta)^\gamma \pi_n(\theta)^{1-\gamma} \big] + p(X|\theta)^\gamma \pi_n(\theta)^{1-\gamma} \ \frac{\partial}{\partial \gamma} \big[ \frac{1}{Z(X;\gamma)} \big] \nonumber \\
&=& \frac{1}{Z(X;\gamma)} \big[ p(X|\theta)^\gamma \pi_n(\theta)^{1-\gamma} \ln \frac{p(X|\theta)}{\pi_n(\theta)} \big] - \frac{p(X|\theta)^\gamma \pi_n(\theta)^{1-\gamma}}{Z(X;\gamma)^2} \ \frac{\partial}{\partial \gamma} \big[ Z(X;\gamma) \big] \nonumber \\
&=& q_\gamma(\theta) \ln \frac{p(X|\theta)}{\pi_n(\theta)} - q_\gamma(\theta) \frac{1}{Z(X;\gamma)} \big[ \int_\theta \ln \frac{p(X|\theta)}{\pi_n(\theta)} \ p(X|\theta)^\gamma \pi_n(\theta)^{1-\gamma} \ d\theta \big] \nonumber \\
&=& q_\gamma(\theta) \ \left( \ln \frac{p(X|\theta)}{\pi_n(\theta)} - \left< \ln \frac{p(X|\theta)}{\pi_n(\theta)} \right>_{q_\gamma} \right). 
\label{eq:DerivativeQGamma} 
\end{eqnarray}
\end{adjustwidth}

Note also that 
\begin{equation}
\int_\theta \frac{\partial}{\partial \gamma} q_\gamma(\theta) \ d\theta \overset{(\ref{eq:DerivativeQGamma})}{=} \int_\theta q_\gamma(\theta) \ \left( \ln \frac{p(X|\theta)}{\pi_n(\theta)} - \left< \ln \frac{p(X|\theta)}{\pi_n(\theta)} \right>_{q_\gamma} \right) \ d\theta = 0.
\label{eq:IntDerEqualZero}
\end{equation}

Then we calculate the derivative of $D_{KL}[q_\gamma||\pi_n]$ with respect to $\gamma$:
\begin{adjustwidth}{-2cm}{}
\begin{eqnarray}
\frac{\partial}{\partial \gamma} D_{KL}[q_\gamma || \pi_n] &=& \int_\theta \frac{\partial}{\partial \gamma} \big[ q_\gamma \ln \frac{q_\gamma(\theta)}{\pi_n(\theta)}\big] \ d\theta \nonumber \\
&=& \int_\theta \left( \ln \frac{q_\gamma(\theta)}{\pi_n(\theta)} \ \frac{\partial}{\partial \gamma} \big[ q_\gamma(\theta) \big] + q_\gamma(\theta) \frac{\partial}{\partial \gamma} \big[ \ln \frac{q_\gamma(\theta)}{\pi_n(\theta)} \big] \right) \ d\theta \nonumber \\
&=& \int_\theta \left( \ln \frac{q_\gamma(\theta)}{\pi_n(\theta)} + 1 \right) \ \frac{\partial}{\partial \gamma} \big[ q_\gamma(\theta) \big]  \ d\theta \nonumber \\
&\overset{(\ref{eq:ModifiedBayesian2})}{=}& \int_\theta \left( \gamma \ln \frac{p(X|\theta)}{\pi_n(\theta)} - \ln Z(X;\gamma) + 1 \right) \ \frac{\partial}{\partial \gamma} \big[ q_\gamma(\theta) \big]  \ d\theta \nonumber \\
&\overset{(\ref{eq:IntDerEqualZero})}{=}& \gamma \int_\theta \left( \ln \frac{p(X|\theta)}{\pi_n(\theta)} \right) \ \frac{\partial}{\partial \gamma} \big[ q_\gamma(\theta) \big]  \ d\theta \nonumber \\
&\overset{(\ref{eq:DerivativeQGamma})}{=}& \gamma \int_\theta \left( \ln \frac{p(X|\theta)}{\pi_n(\theta)} \right) \ \left( \ln \frac{p(X|\theta)}{\pi_n(\theta)} - \left< \ln \frac{p(X|\theta)}{\pi_n(\theta)} \right>_{q_\gamma} \right) q_\gamma(\theta) \ d\theta \nonumber \\
&=& \gamma \int_\theta \left( \ln \frac{p(X|\theta)}{\pi_n(\theta)} \right)^2 q_\gamma(\theta) \ d\theta  \nonumber \\
& & \quad - \ \gamma \left< \ln \frac{p(X|\theta)}{\pi_n(\theta)} \right>_{q_\gamma} \int_\theta \left( \ln \frac{p(X|\theta)}{\pi_n(\theta)} \right) q_\gamma(\theta) \ d\theta \nonumber \\
&=&  \gamma \left( \left< \left( \ln \frac{p(X|\theta)}{\pi_n(\theta)} \right)^2 \right>_{q_\gamma} - \left( \left< \ln \frac{p(X|\theta)}{\pi_n(\theta)} \right>_{q_\gamma} \right)^2 \right) \nonumber \\
&=& \gamma \ var[\ln \frac{p(X|\theta)}{\pi_n(\theta)}] \geq 0.
\end{eqnarray}
\end{adjustwidth}

\subsection{The impact of a data sample $X$ on belief update increases with $\gamma$ in the SMiLe rule}

To prove the statement above we need to show that the impact function $\Delta S_{cc}(X;L)$ in Eq~(\ref{eq:InfluenceFunctionVeryOriginal}), where the SMiLe rule Eq~(\ref{eq:ModifiedBayesian2}) is used for belief update (i.e., when $\pi_{n+1}=q_\gamma$), increases with the parameter $\gamma$. In the following we show that the first derivative of the impact function $\Delta S_{cc}(X;L(\gamma))$ with respect to $\gamma$ is always non-negative.
\begin{adjustwidth}{-2cm}{}
\begin{eqnarray}
\frac{\partial}{\partial \gamma} \Delta S_{cc}(X; L(\gamma)) &=& - \frac{\partial}{\partial \gamma} S_{cc}(X;q_\gamma) \overset{(\ref{eq:AverageShannonModified})}{=} \int_\theta \frac{\partial}{\partial \gamma} \big[ q_\gamma \ln \frac{p(X|\theta)}{q_\gamma(\theta)}\big] \ d\theta \nonumber \\
&=& \int_\theta \left( \ln \frac{p(X|\theta)}{q_\gamma(\theta)} \ \frac{\partial}{\partial \gamma} \big[ q_\gamma(\theta) \big] + q_\gamma(\theta) \frac{\partial}{\partial \gamma} \big[ \ln \frac{p(X|\theta)}{q_\gamma(\theta)} \big] \right) \ d\theta \nonumber \\
&=& \int_\theta \left( \ln \frac{p(X|\theta)}{q_\gamma(\theta)} - 1 \right) \ \frac{\partial}{\partial \gamma} \big[ q_\gamma(\theta) \big]  \ d\theta \nonumber \\
&\overset{(\ref{eq:ModifiedBayesian2})}{=}& \int_\theta \left( (1-\gamma) \ln \frac{p(X|\theta)}{\pi_n(\theta)} + \ln Z(X;\gamma) - 1 \right) \ \frac{\partial}{\partial \gamma} \big[ q_\gamma(\theta) \big]  \ d\theta \nonumber \\
&\overset{(\ref{eq:IntDerEqualZero})}{=}& (1-\gamma) \int_\theta \left( \ln \frac{p(X|\theta)}{\pi_n(\theta)} \right) \frac{\partial}{\partial \gamma} \big[ q_\gamma(\theta) \big] \ d\theta \nonumber \\
&\overset{(\ref{eq:DerivativeQGamma})}{=}& (1-\gamma) \int_\theta \left( \ln \frac{p(X|\theta)}{\pi_n(\theta)} \right) \left( \ln \frac{p(X|\theta)}{\pi_n(\theta)} - \left< \ln \frac{p(X|\theta)}{\pi_n(\theta)} \right>_{q_\gamma} \right) q_\gamma(\theta) \ d\theta \nonumber \\
&=& (1-\gamma) \int_\theta \left( \ln \frac{p(X|\theta)}{\pi_n(\theta)} \right)^2 q_\gamma(\theta) \ d\theta  \nonumber \\
& & \quad - \ (1-\gamma) \left< \ln \frac{p(X|\theta)}{\pi_n(\theta)} \right>_{q_\gamma} \int_\theta \left( \ln \frac{p(X|\theta)}{\pi_n(\theta)} \right) q_\gamma(\theta) \ d\theta \nonumber \\
&=&  (1-\gamma) \left( \left< \left( \ln \frac{p(X|\theta)}{\pi_n(\theta)} \right)^2 \right>_{q_\gamma} - \left( \left< \ln \frac{p(X|\theta)}{\pi_n(\theta)} \right>_{q_\gamma} \right)^2 \right) \nonumber \\
&=& (1-\gamma) \ var[\ln \frac{p(X|\theta)}{\pi_n(\theta)}] \geq 0.
\label{eq:DerivativeDeltaStilde}
\end{eqnarray}
\end{adjustwidth}

\subsection{A larger reduction in the surprise implies a bigger change in belief}

The minimal value of the Lagrangian $\mathbb{L}(q,\lambda)$ in Eq~(\ref{eq:Lagrangian}) that is achieved by the updated belief $q_\gamma$ in Eq~(\ref{eq:ModifiedBayesian2}), obtained by the SMiLe rule, is equal to
\begin{adjustwidth}{-2cm}{}
\begin{eqnarray}
\mathbb{L}(q_\gamma,\lambda) &\overset{(\ref{eq:Lagrangian})}{=}& \left< -\ln p(X|\theta) + \ln q_\gamma(\theta) +\frac{1}{\lambda} \ln \frac{q_\gamma(\theta)}{\pi_n(\theta)} \right>_{q_\gamma} \overbrace{-\frac{B}{\lambda}+\ln ||p||}^{=C} \nonumber \\
&=& \left< -\ln p(X|\theta) + \ln \frac{p(X|\theta)^\gamma \pi_n(\theta)^{1-\gamma}}{Z(X;\gamma)} +\frac{1}{\lambda} \ln \frac{p(X|\theta)^\gamma \pi_n(\theta)^{1-\gamma}}{Z(X;\gamma) \pi_n(\theta)} \right>_{q_\gamma} +C \nonumber \\
&=& \left< (-1+\gamma + \frac{\gamma}{\lambda}) \ln p(X|\theta) + (1-\gamma-\frac{\gamma}{\lambda}) \ln \pi_n - (1+\frac{1}{\lambda}) \ln Z(X;\gamma)\right>_{q_\gamma} +C \nonumber \\
&=& \left< \left(-1+\gamma (1 + \frac{1}{\lambda}) \right) \ln \frac{p(X|\theta)}{\pi_n(\theta)} - (1+\frac{1}{\lambda}) \ln Z(X;\gamma)\right>_{q_\gamma} + C \nonumber \\
&=& - \frac{1}{\gamma} \ln Z(X;\gamma) +C.
\label{eq:MinimumLagrangian2}
\end{eqnarray}
\end{adjustwidth}

Note that we used the equality $\frac{1}{\gamma}=1+\frac{1}{\lambda}$, from Eq~(\ref{eq:GammaFromLambda}), in the last line of Eq~(\ref{eq:MinimumLagrangian2}). If the optimal solution $q_\gamma$ is approximated by any other potential next belief $q$, then its corresponding functional value $\mathbb{L}(q,\lambda)$ differs from its minimal value $\mathbb{L}(q_\gamma,\lambda)$ in proportion to the KL divergence $D_{KL}[ q || q_\gamma]$. This is because,
\begin{adjustwidth}{-2cm}{}
\begin{eqnarray}
\mathbb{L}(q,\lambda)-\mathbb{L}(q_\gamma,\lambda) &\overset{(\ref{eq:Lagrangian}),(\ref{eq:MinimumLagrangian2})}{=}& \left< -\ln p(X|\theta) + \ln q(\theta) + \frac{1}{\lambda} \ln \frac{q(\theta)}{\pi_n(\theta)} \right>_{q} + \frac{1}{\gamma} \ln Z(X;\gamma)\nonumber \\
&=& \frac{1}{\gamma} \left< -\ln p(X|\theta)^\gamma + \ln q(\theta)^\gamma + \ln \left( \frac{q(\theta)}{\pi_n(\theta)} \right)^{\frac{\gamma}{\lambda}}+ \ln Z(X; \gamma) \right>_{q} \nonumber\\
&=& \frac{1}{\gamma} \left< \ln \frac{q(\theta)^{\gamma(1+\frac{1}{\lambda})}Z(X;\gamma)}{p(X|\theta)^\gamma \pi_n(\theta)^{\frac{\gamma}{\lambda}}} \right>_{q} = \frac{1}{\gamma} \left< \ln \frac{q(\theta)Z(X;\gamma)}{p(X|\theta)^\gamma \pi_n(\theta)^{1-\gamma}}\right>_{q} \nonumber \\
&=& \frac{1}{\gamma} D_{KL}[q||q_\gamma].
\label{eq:LagrangianDifference2}
\end{eqnarray}
\end{adjustwidth}

Replacing $q$ with $\pi_n$ in Eq~(\ref{eq:LagrangianDifference2}) follows the impact function $\Delta S_{cc}(X;L)$ in Eq~(\ref{eq:InfluenceFunctionVeryOriginal}) to be,
\begin{eqnarray}
\Delta S_{cc}(X;L(\gamma)) &=& S_{cc}(X;\pi_n) - S_{cc}(X;q_\gamma) \nonumber \\ &\overset{(\ref{eq:Lagrangian})} = & \mathbb{L}(\pi_n,\lambda) + \frac{1}{\lambda} B - \mathbb{L}(q_\gamma,\lambda) - \frac{1}{\lambda} ( B- D_{KL}[q_\gamma|| \pi_n] ) \nonumber \\
& =& \mathbb{L}(\pi_n,\lambda)-\mathbb{L}(q_\gamma,\lambda) + \frac{1}{\lambda} D_{KL}[q_\gamma||\pi_n] \nonumber \\
&\overset{(\ref{eq:LagrangianDifference2})} = & \frac{1}{\gamma} D_{KL}[\pi_n||q_\gamma] + \frac{1}{\lambda} D_{KL}[q_\gamma || \pi_n] \nonumber \\
&\overset{(\ref{eq:GammaFromLambda})} =& \frac{1}{\gamma} D_{KL}[\pi_n||q_\gamma] + \left(\frac{1}{\gamma} -1\right) D_{KL}[q_\gamma ||\pi_n] \geq 0.
\label{eq:ReductionAmount2}
\end{eqnarray}
Note that $q_\gamma$ is the updated belief under the SMiLe rule, i.e., $\pi_{n+1}=q_\gamma$. 
Therefore, the reduction in the  surprise upon a second exposure to the same data sample
is related to the belief changes $D_{KL}[\pi_n||\pi_{n+1}]$ and $D_{KL}[\pi_{n+1} ||\pi_n]$ via Eq~(\ref{eq:ReductionAmount2}). Note that the equality in Eq~(\ref{eq:ReductionAmount2}) holds if and only if there is no change in the current belief, i.e., if $q_\gamma = \pi_{n+1} = \pi_n$. This happens only if $\gamma=0$ which is equivalent to
neglecting the new data point when updating the  belief.

\subsection{The SMiLe rule for beliefs described by Gaussian distribution}

Suppose we have drawn $n-1$ samples $X_1,...,X_{n-1}$ from a Gaussian distribution of known variance $\sigma_x^2$, but unknown mean. The empirical mean after $n-1$ samples is $\hat{\mu}_{n-1}$.  

Assume that the current belief about the mean $\mu$ is a normal distribution, i.e., $\pi_{n-1}(\mu) \sim \mathcal{N}(\hat{\mu}_{n-1}, \sigma_{n-1}^2)$. Since the likelihood of receiving a new sample $X_n$ is also normal, i.e., $p(X_n|\mu) \sim \mathcal{N}(\mu,\sigma_x^2)$, the updated belief obtained by the SMiLe rule [Eq~(\ref{eq:ModifiedBayesian2})] is

\begin{eqnarray}
q_\gamma(\mu) &\propto & \left(exp\left(-\frac{(X_n - \mu)^2}{2\sigma_{x}^2}\right)\right)^\gamma  \left(exp\left(-\frac{(\mu-\hat{\mu}_{n-1})^2}{2\sigma_{n-1}^2}\right)\right)^{1-\gamma} \nonumber \\
&\propto & exp\left(-\frac{(X_n - \mu)^2}{2(\sigma_{x}')^2}\right) \ exp\left(-\frac{(\mu-\hat{\mu}_{n-1})^2}{2(\sigma_{n-1}')^2}\right),
\end{eqnarray}
where $(\sigma_{x}')^2 = \sigma_x^2 / \gamma$ and $(\sigma_{n-1}')^2 = \sigma_{n-1}^2 / (1-\gamma)$. Because the product of two Gaussians is a Gaussian, we arrive at a  distribution $q_\gamma \sim \mathcal{N}(\hat{\mu}_{n}, \sigma_n^2)$ with the mean $\hat{\mu}_n = w_n X_n + (1-w_n)\hat{\mu}_{n-1}$ (with $w_n=\frac{(\sigma_{n-1}')^2}{(\sigma_{x}')^2 + (\sigma_{n-1}')^2}$), and the variance $\sigma_n^2 = \left( \frac{1}{(\sigma_{x}')^2} + \frac{1}{(\sigma_{n-1}')^2} \right)^{-1}$; see~\citep{mackay2003information} for the exact derivation. Assuming $\sigma_{n-1}^2=\sigma_x^2$, then $w_n=\gamma$. Moreover, we can evaluate the confidence-corrected surprise to be

\begin{equation}
S_{cc} (X_n;\pi_{n-1}) = D_{KL}[\mathcal{N}(\hat{\mu}_{n-1},\sigma_{n-1}^2)||\mathcal{N}(X_n, \sigma_x^2)] = \frac{(X_n - \hat{\mu}_{n-1})^2}{2\sigma_x^2}.
\label{eq:surp-Gauss}
\end{equation}

Note that in Eq~(\ref{eq:surp-Gauss}), we used the following equality in Eq~(\ref{eq:KLGauss}) (assuming $\sigma_x^2=\sigma_{n-1}^2$),
 
\begin{equation}
D_{KL}[\mathcal{N}(a_1,b_1^2)||\mathcal{N}(a_2,b_2^2)] = \frac{(a_1-a_2)^2}{2b_2^2} + \frac{1}{2} \left( \frac{b_1^2}{b_2^2} -1 - \ln \frac{b_1^2}{b_2^2} \right).
\label{eq:KLGauss}
\end{equation}

\subsection{The SMiLe rule for beliefs described by a Dirichlet distribution}

Assume that the current belief about the probability of transition from state $s\in \{1,2,...,D\}$ to all $D-1$ possible next states $\check{s}\in \{1,2,...,D\} \backslash s$ is described by a Dirichlet distribution $\pi_n (\mathbf{\theta}_s) \propto \Pi_{\check{s}} \ \theta(s,\check{s})^{\alpha(s,\check{s})-1}$ parametrized by $\mathbf{\alpha}_s = \alpha(s,:)$. Here, $\mathbf{\theta}_s=\theta(s,:)$ denotes a \emph{vector} of random variable $\theta(s,\check{s})$ that determines the probability of transition from $s$ to $\check{s}$, i.e., $0\leq \theta(s,\check{s}) \leq 1$ and $\sum_{\check{s}}\theta(s,\check{s})=1$. The likelihood function for an occurred transition $X:s \rightarrow s'$ is $p(X|\mathbf{\theta}_s)= \theta(s,s')=\Pi_{\check{s}} \ \theta(s,\check{s})^{[\check{s}=s']}$, where $[.]$ denotes the Iverson bracket (that is equal to $1$ if the condition inside the bracket is correct and $0$ otherwise). Therefore, the updated belief $q_\gamma(\mathbf{\theta}_s)$ obtained by the SMiLe rule [Eq~(\ref{eq:ModifiedBayesian2})],

\begin{equation}
q_\gamma(\mathbf{\theta}_s) \propto \left( \Pi_{\check{s}} \ \theta(s,\check{s})^{[\check{s}=s']} \right)^\gamma . \left( \Pi_{\check{s}} \ \theta(s,\check{s})^{\alpha(s,\check{s})-1} \right)^{1-\gamma} \propto \Pi_{\check{s}} \ \theta(s,\check{s})^{\beta(s,\check{s})-1},
\end{equation} 
is again a Dirichlet distribution parametrized by $\beta(s,\check{s}) = (1-\gamma) \alpha(s,\check{s}) + \gamma (1+[\check{s}=s'])$. 

The probability $\hat{T}[t](s,s')$ of transition from $s$ to $s'$ at time step $t$ is estimated by $\hat{T}[t](s,s')=\frac{\alpha[t](s,s')-1+\epsilon}{\sum_{\check{s}} \left(\alpha[t](s,\check{s})-1+\epsilon\right)}$, where $\alpha[t](s,\check{s})$ denotes the updated model parameter at time step $t$. Here, $\epsilon>0$ is a very small number which prevents the denominator from being zero.

\subsection{The online EM algorithm for the maze-exploration task}

The online EM algorithm, presented in~\citep{mongillo2008online}, is an estimation algorithm for the unknown parameters of a hidden Markov model (HMM). For the maze-exploration task we adapted the method presented in~\citep{mongillo2008online} such that the transition probability to a new room also depends on the previously visited room (and not just the current environment). The HMM of the maze-exploration task consists of two sets of unknown parameters: (i) a set ${\bf P}=[P_{ij}]_{2\times 2}$ of (unknown) switch probabilities from environment $i$ to $j$ (where we use $1$ for environment $\mathcal{A}$ and $2$ for environment $\mathcal{B}$), and (ii) a set ${\bf T}=[T_{jss'}]_{2\times 16 \times 16}$ of state transition probabilities, where $T_{jss'}$ denotes the probability of transition from state $s$ to state $s'$ within environment $j$. The set of all unknown parameters is denoted by ${\bf \theta} \equiv ({\bf P}, {\bf T})$.

At each time step $t$, we estimate the probability $q_l^{t}=P(E_t=l|s_{0\rightarrow t})$ of being in environment $E_t=l\in\{1,2\}$, given all previous state transitions $s_{0\rightarrow t}=\{s_0,s_1,...,s_t\}$. The probability $q_l^{t}$ can be recursively calculated by
  
\begin{equation}
\hat{q}_l^{t} = \sum_m \hat{q}_m^{t-1} \gamma_{ml}^t,
\end{equation}
where $\gamma_{ml}^t =\frac{P(s'=s_t|s=s_{t-1},E_t=l)P(E_t=l|E_{t-1}=m)}{P(s'=s_t|s_{0\rightarrow (t-1)})}$ belongs to a set of auxiliary variables ${\bf \Gamma}=[\gamma_{lh}]_{2\times 2}$ that are calculated by the last estimate $\hat{{\bf \theta}}^{t-1}$ of the model parameters:

\begin{equation}
\gamma_{lh}^{t} = \frac{\hat{P}_{lh}^{t-1} \hat{T}_{hs_{t-1}s_{t}}^{t-1}}{\sum_{m,n} \hat{q}_m^{t-1} \hat{P}_{mn}^{t-1}\hat{T}_{ns_{t-1}s_{t}}^{t-1}}.
\end{equation}

Then, using these auxiliary variables $\gamma_{lh}$, a set ${\bf \Phi}=[\hat{\phi}_{i,j,s,s',h}]_{2\times 2\times 16\times 16\times 2}$ of parameters is recursively updated:
 
\begin{equation}
\hat{\phi}_{i,j,s,s',h}^t= \sum_l \gamma_{lh}^{t} \Big[(1-\eta) \hat{\phi}_{i,j,s,s',l}^{t-1} + \eta \hat{q}_l^{t-1} \Delta_{ijss'}^{lhs_{t-1}s_t}\Big],
\end{equation}
where $\Delta_{ijss'}^{lhs_{t-1}s_t} = \delta(i-l) \delta(j-h) \delta(s-s_{t-1}) \delta(s'-s_{t})$, $\delta(.)$ is the Kronecker delta (i.e., $1$ when its argument is zero and $0$ otherwise), and $\eta$ is the learning rate.

Finally, the model parameters are updated by 

\begin{equation}
\hat{P}_{ij}^t = \frac{\sum_{s,s',h} \hat{\phi}_{ijss'h}^{t}}{\sum_{j,s,s',h} \hat{\phi}_{ijss'h}^{t}}; \quad \hat{T}_{jss'}^{t} = \frac{\sum_{i,h} \hat{\phi}_{ijss'h}^{t}}{\sum_{i,s',h} \hat{\phi}_{ijss'h}^{t}}.
\end{equation}

We emphasize that in order for the online EM algorithm to work properly, some technical considerations must be respected. For instance, in the beginning of learning, only online estimation of ${\bf \Phi}$ must be updated (without updating the model parameters ${\bf \theta}$), so that the estimation error for the first $2000$ time steps of our simulation (Fig~\ref{fig:Fig 6}A, blue) remains fixed. Moreover, we found that the online EM algorithm works well only if it is correctly initialized. To make our comparison fair, we assumed the agent ``believes in'' frequent transitions between environments by initializing the probabilities $\hat{P}_{ij}^0$ that describe the switch between environment $\mathcal{A}$ and $\mathcal{B}$ to be very close to true ones. Without such an assumption, the online EM takes even more time than what we reported here to learn the maze-exploration task. The actual initialization values were $\hat{P}_{12}^0=\hat{P}_{21}^0=0.1$ while the true values were $P_{12}=P_{21}=0.005$.

\section*{Acknowledgments}
This project has been funded by the European Research Council under grant agreement No. 268689, and by the European Union’s Horizon 2020 research and innovation program under grant agreement No. 720270.

\nolinenumbers

\bibliographystyle{frontiersinSCNS_ENG_HUMS}
\bibliography{mybib}

\end{document}